%% file: main.tex
\PassOptionsToPackage{table,xcdraw}{xcolor}
\PassOptionsToPackage{most}{tcolorbox} 
\documentclass[11pt, a4paper, logo, copyright, nonumbering]{wechat}
\usepackage[numbers, square, sort&compress]{natbib}
\usepackage{dblfloatfix}
\usepackage{ulem}
\usepackage{caption}
\definecolor{citecolor}{HTML}{1976D2}
\hypersetup{
    colorlinks=true,
    linkcolor=black,
    filecolor=magenta,
    urlcolor=blue!50!black,
    citecolor=citecolor,
}
\usepackage{dramatist}
\usepackage{xspace}
\usepackage{pifont}
\usepackage{multirow}
\usepackage{tcolorbox}
\usepackage{xltabular}
\usepackage{longtable}
\usepackage{hyperref}
\usepackage{diagbox}
\usepackage{makecell}
\usepackage{amsmath}
\usepackage{amssymb}
\usepackage{amsfonts}
\usepackage{bm}
\usepackage{lineno}

\usepackage[bottom]{footmisc}

\usepackage{CJKutf8}

\usepackage{setspace}
\usepackage{subcaption} 

\usepackage{titlesec}

\usepackage{tcolorbox}
\usepackage{fontawesome5}
\usepackage{enumitem}

\usepackage{booktabs}
\usepackage{multirow}
\usepackage[table,xcdraw]{xcolor}
\usepackage{graphicx}

\usepackage{algorithm}
\usepackage{algorithmic}
\usepackage{subcaption}
\usepackage{microtype}
\usepackage{mathtools}
\usepackage{array}
\usepackage{tabularx}
\usepackage{colortbl}
\usepackage{listings}

\usepackage{tikz}
\usetikzlibrary{arrows.meta,positioning,fit,backgrounds}

\usepackage[nameinlink,capitalize,noabbrev]{cleveref}

\tcbset{
    contributionbox/.style={
        colback=blue!3,
        colframe=blue!50!black,
        boxrule=0.5pt,
        arc=3pt,
        left=6pt,
        right=6pt,
        top=6pt,
        bottom=6pt,
        fonttitle=\bfseries
    }
}

\titlespacing*{\paragraph}
  {0pt}                   
  {0.5ex plus 1ex minus .2ex} 
  {1em}

\makeatletter
\def\@BTrule[#1]{%
  \ifx\longtable\undefined
    \let\@BTswitch\@BTnormal
  \else\ifx\hline\LT@hline
    \nobreak
    \let\@BTswitch\@BLTrule
  \else
     \let\@BTswitch\@BTnormal
  \fi\fi
  \global\@thisrulewidth=#1\relax
  \ifnum\@thisruleclass=\tw@\vskip\@aboverulesep\else
  \ifnum\@lastruleclass=\z@\vskip\@aboverulesep\else
  \ifnum\@lastruleclass=\@ne\vskip\doublerulesep\fi\fi\fi
  \@BTswitch}
\makeatother

\addto\extrasenglish{
}

 {\begin{list}{}%
         {\setlength{\leftmargin}{#1}}%
         \item[]%
 }
 {\end{list}}

\reportnumber{001} 

\colorlet{failred}{red!75!black}
\colorlet{okgreen}{green!50!black}
\colorlet{obsgray}{black!55}
\newcommand{\hard}[1]{\textcolor{failred}{#1}}   
\newcommand{\good}[1]{\textcolor{okgreen}{#1}}   
\newcommand{\ttag}[1]{\textcolor{black!70}{\textbf{#1}}}   
\newcommand{\obs}[1]{\textcolor{obsgray}{#1}}    
\newtcolorbox{casebox}[1]{%
  enhanced, breakable, colback=white, colframe=black!55,
  boxrule=0.5pt, arc=1.2mm, left=2.2mm, right=2.2mm, top=1.8mm, bottom=1.8mm,
  title={#1}, coltitle=white, colbacktitle=black!68,
  fonttitle=\footnotesize\bfseries, before skip=8pt, after skip=8pt}
\newtcolorbox{basepanel}{%
  colback=failred!5, colframe=failred!80, boxrule=0.5pt, arc=1mm,
  left=1.6mm, right=1.6mm, top=1.2mm, bottom=1.2mm,
  title={$\times$~\textbf{Base} (Qwen3.5-9B)}, coltitle=white, colbacktitle=failred!85,
  fonttitle=\scriptsize\bfseries}
\newtcolorbox{ourspanel}{%
  colback=okgreen!6, colframe=okgreen!85, boxrule=0.5pt, arc=1mm,
  left=1.6mm, right=1.6mm, top=1.2mm, bottom=1.2mm,
  title={\textbf{Pass: SkillRubric (Ours)}}, coltitle=white, colbacktitle=okgreen!90,
  fonttitle=\scriptsize\bfseries}

\usepackage[most]{tcolorbox}
\usepackage{listings}
\tcbuselibrary{listings,breakable,skins}

\definecolor{prompttitle}{HTML}{287780}
\definecolor{promptborder}{HTML}{A9C8CC}
\definecolor{promptbackground}{HTML}{F7F9F9}

\lstdefinestyle{promptstyle}{
  basicstyle=\small\ttfamily,
  columns=fullflexible,
  keepspaces=true,
  showspaces=false,
  showstringspaces=false,
  breaklines=true,
  breakatwhitespace=true,
  breakautoindent=false,
  breakindent=0pt,
  xleftmargin=0pt,
  framexleftmargin=0pt,
  aboveskip=0pt,
  belowskip=0pt,
  escapeinside={(*@}{@*)}
}

\newtcblisting{promptbox}[1]{%
  enhanced,
  breakable,
  listing only,
  listing engine=listings,
  colback=promptbackground,
  colframe=promptborder,
  boxrule=0.6pt,
  arc=0mm,
  left=4mm,
  right=4mm,
  top=8mm,
  bottom=3mm,
  title={#1},
  colbacktitle=prompttitle,
  coltitle=white,
  fonttitle=\bfseries\ttfamily,
  attach boxed title to top left={xshift=4mm,yshift=-3mm},
  boxed title style={%
    colback=prompttitle,
    colframe=prompttitle,
    boxrule=0pt,
    arc=1mm,
    left=2mm,
    right=2mm,
    top=1mm,
    bottom=1mm
  },
  listing options={style=promptstyle},
  before skip=8pt,
  after skip=8pt
}

\colorlet{guideblue}{blue!65!black}
\colorlet{rubricorange}{orange!85!black}

\newtcolorbox{skillbox}[1]{%
  enhanced,
  colback=white,
  colframe=black!55,
  boxrule=0.5pt,
  arc=1.2mm,
  left=2.2mm,
  right=2.2mm,
  top=1.8mm,
  bottom=1.8mm,
  title={#1},
  coltitle=white,
  colbacktitle=black!68,
  fonttitle=\footnotesize\bfseries,
  fontupper=\footnotesize,
  before upper={%
    \setlength{\parindent}{0pt}%
    \setlength{\parskip}{1.0mm}%
  },
  before skip=0pt,
  after skip=2pt
}

\newtcolorbox{guidancepanel}{%
  enhanced,
  colback=guideblue!5,
  colframe=guideblue!80,
  boxrule=0.5pt,
  arc=1mm,
  left=1.6mm,
  right=1.6mm,
  top=1.2mm,
  bottom=1.2mm,
  title={Actor-facing guidance $g$ (shown to the policy)},
  coltitle=white,
  colbacktitle=guideblue!85,
  fonttitle=\scriptsize\bfseries,
  fontupper=\scriptsize,
  before upper={%
    \setlength{\parindent}{0pt}%
    \setlength{\parskip}{0.9mm}%
  },
  before skip=0pt,
  after skip=0pt
}

\newtcolorbox{rubricpanel}{%
  enhanced,
  colback=rubricorange!6,
  colframe=rubricorange!85,
  boxrule=0.5pt,
  arc=1mm,
  left=1.6mm,
  right=1.6mm,
  top=1.2mm,
  bottom=1.2mm,
  title={Evaluator-facing rubric $\mathcal{C}$ (hidden from the policy)},
  coltitle=white,
  colbacktitle=rubricorange!90,
  fonttitle=\scriptsize\bfseries,
  fontupper=\scriptsize,
  before upper={%
    \setlength{\parindent}{0pt}%
    \setlength{\parskip}{0.9mm}%
  },
  before skip=0pt,
  after skip=0pt
}

\newcounter{srprompt}[section]
\renewcommand{\thesrprompt}{\thesection.\arabic{srprompt}}

\crefname{srprompt}{Prompt}{Prompts}
\Crefname{srprompt}{Prompt}{Prompts}

\newtcblisting{skillrubricprompt}[1]{%
    enhanced,
    breakable,
    listing only,
    colback=gray!3,
    colframe=gray!45,
    boxrule=0.5pt,
    arc=1pt,
    left=1.5mm,
    right=1.5mm,
    top=1mm,
    bottom=1mm,
    title={#1},
    fonttitle=\bfseries\small,
    listing options={
        basicstyle=\ttfamily\scriptsize,
        breaklines=true,
        breakatwhitespace=false,
        columns=fullflexible,
        keepspaces=true,
        showstringspaces=false,
        tabsize=2
    }
}

\title{\centering SkillRubric: Co-Evolving Actor Guidance and Evaluator Rubrics for Multimodal Agents}

\author{
    {\large\bfseries
    Bingqing Jiang$^{1,3,\ast,\S}$,
    Guoxi Zhang$^{2,\ast}$,
    Jasper Wang$^{3}$,
    Auric Wang$^{3}$,
    Bingning Wang$^{3,\ddagger}$,
    Tianyi Lin$^{4}$,
    Zichao Yu$^{1}$,
    Yujin Han$^{1}$,
    Ziye Ma$^{5}$,
    Difan Zou$^{1,\dagger}$
    }\\
    {\normalsize\normalfont
    $^{1}$The University of Hong Kong
    \quad
    $^{2}$Peking University
    \quad
    $^{3}$WeChat, Tencent
    \quad
    $^{4}$The Hong Kong Polytechnic University
    \quad
    $^{5}$City University of Hong Kong
    }
}

\input{commands.tex}

\begin{abstract}
Recent work incorporates reusable skills distilled from past interactions into multimodal agent training, providing procedural guidance for long-horizon planning and tool use. However, policy optimization in these methods remains driven primarily by sparse outcome rewards, providing little supervision for intermediate decisions. Rubric-based rewards address this limitation through explicit intermediate criteria, but reliable rubrics are difficult to construct at scale and often disconnected from the procedure followed by the actor. We observe that a well-structured skill naturally specifies both how to act and what successful execution should achieve. Based on this insight, we introduce \textbf{SkillRubric}, which represents each skill through aligned actor-facing guidance and an evaluator-facing rubric. A multimodal verifier evaluates skill-defined goals using screenshots and tool outputs, assigning completion and progress rewards to the responsible turns. We further introduce an alternating co-evolution scheme that validates guidance revisions through paired rollouts under a frozen policy and rubric revisions offline under fixed guidance. Experiments across diverse multimodal agent benchmarks demonstrate consistent performance gains, while controlled paired rollouts further show that evolved skills provide more effective guidance for planning and tool use than their preceding versions.
\end{abstract}

\begin{document}
\begin{CJK*}{UTF8}{gbsn}

\maketitle
\setlength{\parindent}{0pt} 
\enlargethispage{1cm}

\newcommand\blfootnote[1]{%
  \begingroup
  \renewcommand\thefootnote{}\footnote{#1}%
  \addtocounter{footnote}{-1}%
  \endgroup
}


\blfootnote{$^\ast$ These authors contributed equally to this work.}
\blfootnote{$^\dagger$ Corresponding author.}
\blfootnote{$^\ddagger$ Project leader.}
\blfootnote{$^\S$ Work done during an internship at WeChat.}

\section{Introduction}
\label{sec:introduction}

Multimodal agents solve complex visual reasoning tasks by repeatedly invoking external tools, such as image editors, code interpreters, and search engines, and incorporating the resulting observations into subsequent decisions. The resulting trajectories extend across many interdependent steps, placing substantial demands on task planning and flexible tool orchestration \citep{guo2025visualtoolbench,li2025tirbench,wu2025mmsearchr1}. Current agents have limited mechanisms for converting prior trajectories into reusable procedural knowledge and therefore often approach each new task largely from scratch. Recent work addresses this problem by distilling past interactions into skills and experience-based guidance that structure planning and tool use on future tasks.
A skill is a compact representation of procedural knowledge that specifies when a strategy applies and how it should be executed through planning and tool use. 
One line of work retrieves skills or experience-derived guidance at inference time to improve planning and tool use without updating model parameters \citep{jiang2026xskill,zhang2026mmskills}. Skill-augmented reinforcement learning extends this paradigm by integrating skill discovery and refinement into policy optimization, allowing the skill library to evolve alongside the policy \citep{xia2026skillrl,he2026reskill}. Despite this tighter integration, policy updates remain driven primarily by sparse outcome rewards, which provide no direct supervision over intermediate decisions \citep{zeng2025turncredit,feng2025gigpo}.

To address this limitation, rubric-based rewards decompose task success into explicit criteria and use a judge to evaluate a trajectory or its intermediate components against these criteria \citep{gunjal2025rubrics,xie2026srar,tian2026arco,li2026rubricem}. By assessing verifiable intermediate objectives, rubrics can distinguish partial progress from complete failure among trajectories with the same unsuccessful outcome, providing more informative supervision than outcome rewards alone. The reliability of this supervision, however, depends critically on rubric quality. Expert-authored rubrics are costly and difficult to scale, whereas automatically generated rubrics rely on the generator to identify appropriate intermediate objectives and evidence requirements, which is particularly difficult for multimodal tool-use trajectories. Consequently, obtaining rubrics that are both scalable and aligned with the agent’s execution procedure remains challenging.

\begin{figure*}[t]
  \centering
  \includegraphics[width=\textwidth]{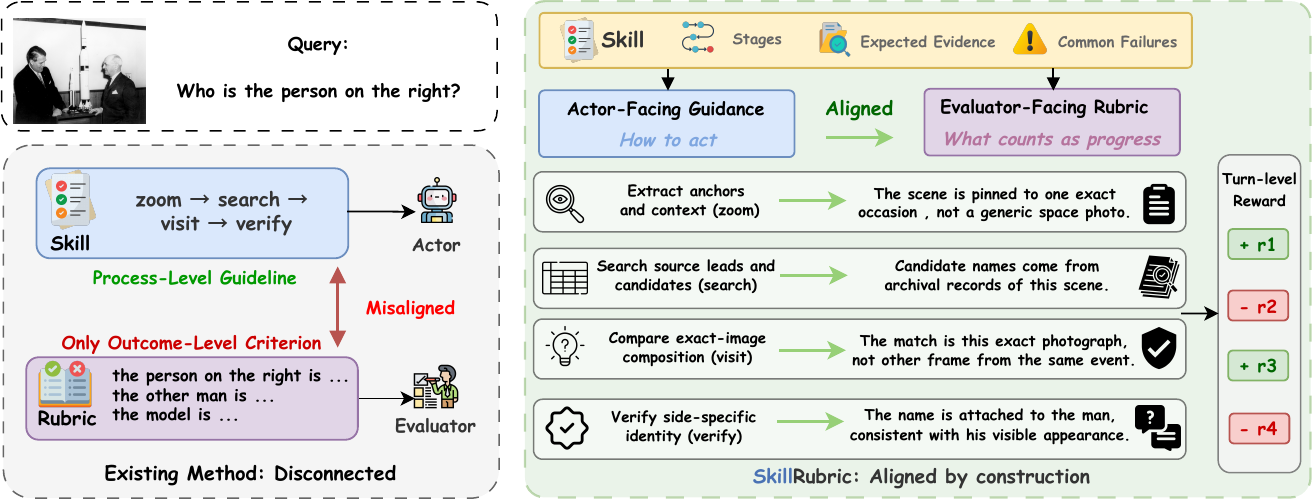}
  \vspace{-5mm}
  \caption{\textbf{Aligned actor guidance and evaluator rubrics through a shared skill representation.}
SkillRubric derives both views from a shared procedural structure, aligning guidance stages with observable evaluation criteria.}
  \label{fig:intro}
  \vspace{-5mm}
\end{figure*}

In seeking a principled source for reliable rubrics, we identify an overlooked structural connection to the skills already used to guide agents. A well-structured skill not only tells an agent how to act, but its procedural structure also naturally defines what should be evaluated and how.
Specifically, as illustrated in \cref{fig:intro}, its stages identify intermediate objectives, its quality criteria and expected evidence characterize successful completion, and its common mistakes specify failure conditions. This observation motivates representing each skill through two aligned views of the same procedural strategy: actor-facing \emph{guidance} that specifies how to act and an evaluator-facing \emph{rubric} that specifies how successful execution should be assessed. Deriving both views from a shared procedural source aligns task execution and process evaluation by construction.

Based on this insight, we introduce \textbf{SkillRubric}, a framework that unifies skill-guided execution and rubric-based policy learning in multimodal agents. SkillRubric operationalizes this insight through dual-view skills: actor-facing guidance directs trajectory generation, while an evaluator-facing rubric derived from the same strategy defines weighted, verifiable intermediate goals. Using screenshots and tool outputs as evidence, a multimodal verifier assesses goal completion and progress and attributes them to the responsible turns, thereby converting skill-derived rubrics into evidence-grounded turn-level rewards. As policy training progresses, residual failures expose deficiencies in both the guidance followed by the agent and the rubric used to evaluate it. SkillRubric therefore introduces an alternating co-evolution scheme that selects guidance revisions through paired rollouts under a frozen policy and validates rubric revisions offline under fixed guidance, enabling both components to improve without conflating changes in agent behavior with changes in the reward specification. Extensive experiments across diverse multimodal agent benchmarks demonstrate consistent performance improvements, with further analyses validating evidence-grounded turn attribution, cost-aware multimodal verification, and alternating co-evolution.
Our contributions are summarized as follows:
\vspace{-2mm}
\begin{itemize}[leftmargin=*]
\item We introduce \textbf{SkillRubric}, which represents each reusable skill through two aligned views: actor-facing guidance for generating trajectories and an evaluator-facing rubric for assessing them. By grounding skill-defined goals in screenshots and tool outputs, SkillRubric turns the same procedural knowledge that guides the agent into evidence-grounded turn-level rewards.
\item We propose a controlled alternating co-evolution scheme that selects guidance updates through paired rollouts under a frozen policy and rubric updates through offline validation under fixed guidance. This separation allows both components to improve without confounding changes in agent behavior and reward specification.
\item Extensive experiments across diverse multimodal agent benchmarks demonstrate consistent improvements in agent performance. Controlled paired rollouts under a frozen policy further demonstrate that evolved skills provide more effective reusable guidance for task planning and tool use than their preceding versions.

\end{itemize}

    
    

\vspace{-2mm}
\section{Related Work}
\label{sec:related_work}
\vspace{-2mm}

\paragraph{Multimodal agents and reusable procedural knowledge.}
Recent work distills interaction trajectories into reusable skills for planning and tool use. Inference-time methods retrieve textual or multimodal procedures from prior experience, whereas reinforcement-learning methods integrate skill discovery and refinement into policy optimization \citep{jiang2026xskill,xie2025mirage,zhang2026mmskills,chen2026cuaskill,xia2026skillrl,li2026skillgraph,he2026reskill}. These approaches primarily use skills for action generation, leaving their role in reward specification largely unexplored; SkillRubric derives evaluator-facing rubrics from the same procedural knowledge.

\vspace{-2mm}
\paragraph{Fine-grained supervision for agentic reinforcement learning.}
Fine-grained agentic supervision has been studied through turn-level advantage estimation, progress-based rewards, and self-distillation. Existing methods estimate local advantages from intermediate rewards or repeated states, derive progress signals from rollout structures or learned potentials, and transfer token-level guidance through privileged-context teachers \citep{zeng2025turncredit,feng2025gigpo,wang2026rtmc,xi2025agentprm,tao2026trace,lu2026sdar,wang2026skillsd}. These approaches infer local utility without explicit semantic targets, whereas SkillRubric defines verifiable process goals from reusable skills.

\vspace{-2mm}
\paragraph{Rubric-based rewards and adaptive evaluation.}
Rubric-based research has advanced along three dimensions: scalable construction, fine-grained attribution, and adaptive evaluation. Existing methods synthesize instance-specific criteria, attribute them to reasoning steps or stages, and adapt rubric generation during policy training \citep{gunjal2025rubrics,liu2025openrubrics,li2026ares,xie2026srar,li2026rubricem,tian2026arco,ding2026evorubrics}. These approaches generally construct rubrics independently of reusable actor skills; SkillRubric derives guidance and evaluation criteria from a shared skill representation and validates their updates separately.

\vspace{-2mm}
\section{Methodology}
\label{sec:method}
\vspace{-2mm}

\begin{figure*}[t]
  \centering
  \includegraphics[width=\columnwidth]{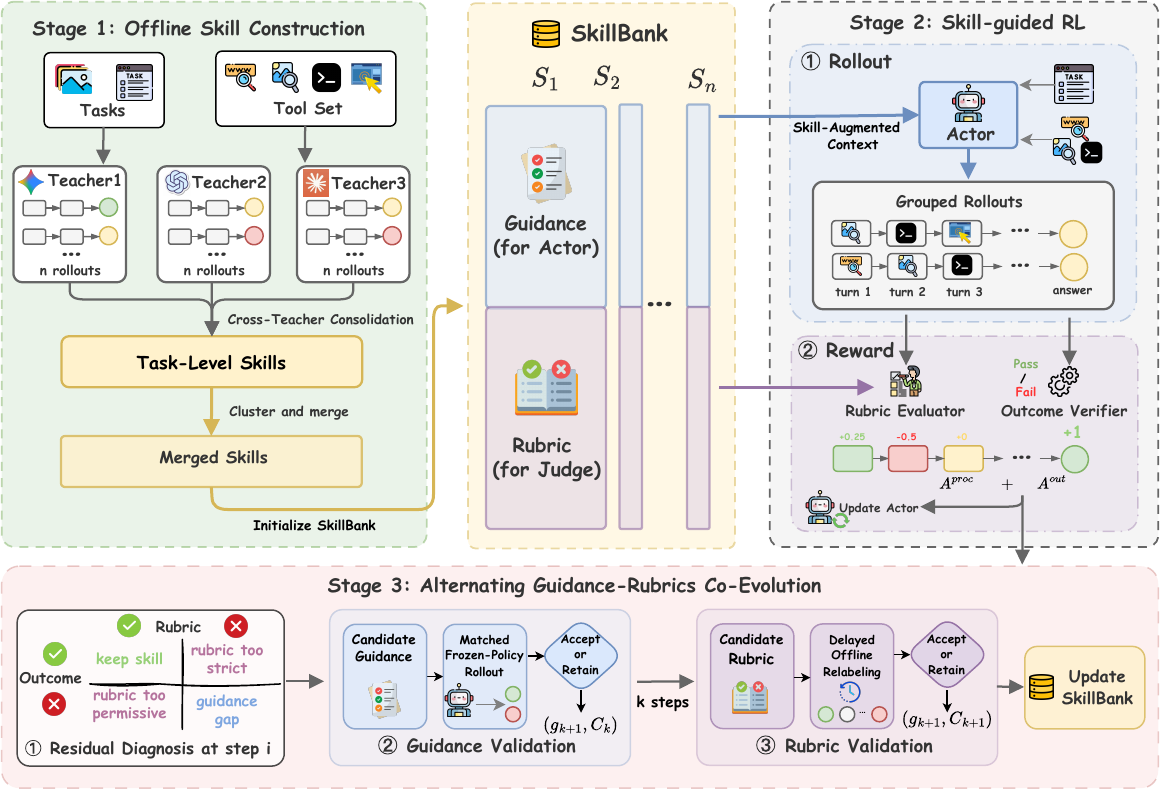}
  \caption{\textbf{Overview of SkillRubric.} \textbf{Stage 1:} Success-filtered multi-teacher trajectories initialize the policy, while all collected trajectories are distilled into a dual-view SkillBank. \textbf{Stage 2:} Actor-facing guidance conditions grouped rollouts, and evaluator-facing rubrics provide process credit alongside a fixed outcome signal. \textbf{Stage 3:} Guidance and rubrics are alternately evolved through outcome-grounded validation, with rubric validation delayed by $H$ policy updates.}\vspace{-5mm}
  \label{fig:method_overview}
\end{figure*}

In this section, we present SkillRubric, which pairs actor-facing guidance with
an evaluator-facing rubric to use each skill for both policy guidance and
process supervision without exposing evaluator information to the actor. As
shown in \cref{fig:method_overview}, Stage 1 initializes the policy from
successful multi-teacher trajectories and distills all trajectories into a
dual-view SkillBank (\cref{sec:skill_initialization}). Stage 2 uses retrieved
guidance for grouped rollouts and rubric verification for outcome-anchored,
responsibility-aware process credit (\cref{sec:policy_optimization}). Stage 3
alternately validates guidance through frozen-policy rollouts and rubrics
through delayed offline relabeling after $H$ policy updates
(\cref{sec:coevolution}). 

\vspace{-2mm}
\subsection{Dual-View SkillBank Construction from Multi-Teacher Trajectories}
\label{sec:skill_initialization}
\vspace{-2mm}




SkillRubric first distills high-quality skills from multi-teacher trajectories, including both successes and failures. All are used for distillation, while successes additionally warm-start the student via SFT. To handle noise and cross-teacher biases, we apply a two-level hierarchical distillation mechanism, denoted by $\mathcal M_{\mathrm{skill}}$: (1) \emph{intra-teacher aggregation}, which extracts task-specific candidate skills by contrasting successful and failed rollouts per teacher; and (2) \emph{cross-teacher consolidation}, which merges these candidates into a unified skill set by reconciling outcome-based conflicts, preserving complementarity, and pruning redundancy. This two-tier design reduces both rollout-level noise and teacher-specific bias, yielding more robust skills than single-stage distillation.

Prior work typically represents such procedural knowledge through an applicability description and actor-facing guidance, using it primarily to improve trajectory generation while overlooking its potential for process supervision~\citep{jiang2026xskill,zhang2026mmskills,xia2026skillrl}. We instead represent each distilled skill as 
\begin{align*}
S=(d,g,\mathcal{C}), \quad d: \text{description},\ g:\text{guidance},\ \mathcal C:\text{set of rubrics}.
\end{align*}
More specifically, the description $d$ identifies the applicability of the skill, the guidance $g$ provides reusable approaches for policy execution, and the rubrics in $\mathcal{C}$ define evaluator-facing criteria for assessing intermediate progress. The paired guidance and rubric are jointly derived by the $\mathcal{M}_{\mathrm{skill}}$ from the distilled procedural abstraction:
$g$ summarizes how to execute the procedure, while $\mathcal{C}$ translates its critical effects into observable progress criteria. 
Importantly, to preserve transferability and generalization, $\mathcal{M}_{\mathrm{skill}}$
removes task-specific answers and fixed execution paths from $g$, and excludes action- or tool-specific requirements from $\mathcal{C}$.
For instance, in visual entity identification, rather than instructing the policy to search for a particular candidate or evaluating whether it performs a specific search or zoom action, $g$ recommends comparing discriminative visual attributes against independent evidence, while $\mathcal{C}$ assesses whether the evidence
consistently supports the identification.

Besides, tasks with different surface forms often rely on the same underlying procedure, so retaining each task-level skill independently would fragment reusable knowledge and limit its transfer across tasks. 
To obtain more general skills, $\mathcal{M}_{\mathrm{skill}}$ directly merges similar task-level skills according to their applicability, procedural guidance, and evaluation criteria. During merging, it consolidates shared principles, deduplicates overlapping instructions, and removes residual task-specific details. The merged entries form the SkillBank $\mathcal{B}$, where each skill identity links an applicability description $d$ with its aligned guidance $g$ and rubric $\mathcal{C}$.
All prompt details and examples are provided in Appendices~\ref{app:prompt_design} and \ref{app:skill_representation}.

\vspace{-2mm}
\subsection{Skill-Conditioned Policy Optimization with Turn-Level Rubrics}
\label{sec:policy_optimization}
\vspace{-2mm}








Having constructed the SkillBank $\mathcal{B}$, we now describe how to employ it for skill-guided reinforcement learning on a target task. A skill $S=(d,g,\mathcal{C})$ is retrieved by matching its applicability $d$ to the task description; its guidance $g$ then conditions policy rollouts, while its rubric $\mathcal{C}$ provides turn-level process supervision. The resulting process advantage, combined with a global outcome advantage, forms the final learning signal for policy optimization.

\textbf{Task-specific Skill Selection.} To enable a skill-guided RL method for a target task, the first step is to retrieve the most appropriate skill from the SkillBank $\mathcal{B}$. We accomplish this by computing cosine similarities between the task description and each skill's applicability description $d$ in a shared embedding space, and then selecting the skill with the highest similarity. Formally, for a skill $S=(d,g,\mathcal{C})$ in the bank and a target task described by $q$, the selected skill is given by $S^* = \arg\max_{S\in\mathcal B}\cos (f_{\mathrm{emb}}(q), f_{\mathrm{emb}}(d))$, where $f_{\mathrm{emb}}(\cdot)$ denotes the embedding function that maps textual descriptions into vector representations.

\textbf{Turn-level Rubric-based Evaluation.} The next step is to leverage the rubrics \(\mathcal C=\{c_j\}_{j=1,\dots,J}\) for assigning process credit to the student roll-out trajectories $\{\tau_i\}_{i=1,\dots,G}$, where $i$ denotes the roll-out index. The student trajectory typically follows a multi-turn structure, where each turn constitutes one trial involving specific tool use. Instead of using standard token-level rewards \citep{cui2025prime,xie2025capo,xu2026rubricstotokens}, we propose turn-level rubric evaluation. Under this scheme, the credit assigned to each token within a turn is determined by the outcome of the entire turn. More specifically, since the effectiveness of an action can often only be evaluated from the returned screenshot or tool output, the turn-level score is evaluated based on both the generated response \(a_t\) and the resulting observation \(o_t\), where \(t\) indexes the turn. This formulation raises two critical challenges: (1) \emph{how should we assign each rubric \(c_j \in \mathcal C\) to its corresponding turn?} and (2)\emph{ since turns are not isolated, how can we effectively incorporate preceding evidence from earlier turns to ensure accurate evaluation?}

To address these challenges, we design a coarse-to-fine rubric verifier. 
At the coarse stage, a lightweight matcher examines the accumulated evidence available by the end of turn $t$, including the current response and any resulting tool observation, and selects a subset
of rubric criteria $\mathcal{R}_{i,t}\subseteq\mathcal{C}$ relevant to that turn.
This context-aware selection avoids evaluating every turn against all criteria, while retaining the historical evidence needed to interpret the current interaction. At the fine stage, for each criterion $c \in \mathcal{C}$, we collect all turns from trajectory $i$ that were matched to it, forming the candidate set $\mathcal{K}_{i,c} = \{ t \mid c \in \mathcal{R}_{i,t} \}$. 
A multimodal verifier then reviews these candidate turns together with their context summaries and observations. 
For each applicable and verifiable criterion, it assigns a binary trajectory-level satisfaction score $s_{i,c}\in\{0,1\}$ indicating whether trajectory $i$ satisfies criterion $c$, together with
non-negative responsibility scores $\alpha_{i,c,t}$ identifying the candidate turns responsible for its satisfaction or failure.

To make these scores comparable across different criteria and trajectories, we apply two separate normalizations. For each criterion $c$, we normalize the responsibility scores $\alpha_{i,c,t}$ across its candidate turns $\mathcal{K}_{i,c}$ to obtain $\bar{\alpha}_{i,c,t}$, and normalize the satisfaction scores $s_{i,c}$ across the $G$ rollouts for each criterion to obtain $\widehat{s}_{i,c}$, i.e.,
\begin{equation*}
\bar{\alpha}_{i,c,t}
=
\frac{
\alpha_{i,c,t}
}{
\sum_{t'\in\mathcal{K}_{i,c}}\alpha_{i,c,t'}
},
\qquad
\widehat{s}_{i,c}
=
\frac{
s_{i,c}-\mu_c
}{
\sigma_c+\varepsilon
},
\label{eq:normalized_rubric_signals}
\end{equation*}
where $\mu_c$ and $\sigma_c$ are the group mean and standard deviation for criterion $c$.

Beyond final contributions, we also measure each turn's incremental progress toward each criterion. For a criterion $c$, we define $\phi_{i,c}(\tau_{i,1:t})$ as the degree to which the accumulated evidence up to turn $t$ supports criterion $c$. The incremental progress of turn $t$ is defined as $
\Delta\phi_{i,c,t} = \phi_{i,c}(\tau_{i,1:t}) - \phi_{i,c}(\tau_{i,1:t-1})$,
where positive, negative, and zero values respectively indicate progress, regression, or no detectable change. Combining these quantities, we are able to obtain the turn-level process advantage as follows: 
\begin{equation*}
A_{i,t}^{\mathrm{proc}} = \operatorname{clip}\bigg(\sum_{c \in \mathcal{R}_{i,t}} w_c \left(\bar{\alpha}_{i,c,t}\widehat{s}_{i,c} + \omega\Delta\phi_{i,c,t}\right), -\delta, \delta\bigg),
\label{eq:process_advantage}
\end{equation*}
where $w_c \in \mathbb{R}_{>0}$ is a predefined weight for criterion $c$, $\omega \in \mathbb{R}_{>0}$ rescales the incremental term to match the numerical magnitude of the responsibility-weighted satisfaction term, and $\delta > 0$ bounds the final advantage to prevent extreme values. The first term attributes the final rubric result to its responsible turns based on their normalized contributions, while the second term explicitly rewards progress or penalizes regression at the turn where it occurs.

\textbf{Policy Objective.} Finally, we combine the turn-level process
advantage with the group-relative outcome advantage
$A_i^{\mathrm{out}}$ \citep{shao2024deepseekmath}, which is computed from
terminal rewards independently of the rubrics $\mathcal{C}$:
\begin{equation*}
A_{i,t} = A_i^{\mathrm{out}} + A_{i,t}^{\mathrm{proc}}.
\label{eq:turn_advantage}
\end{equation*}
All policy-generated tokens within the same assistant turn $t$ share this advantage $A_{i,t}$. Environment-generated content such as screenshots and tool outputs are excluded from the policy loss. 
The policy is then optimized via the clipped PPO objective with KL regularization \citep{schulman2017proximal,ouyang2022training}:
\begin{equation*}
\mathcal{L}(\theta) = -\mathbb{E}_{i,t,k}\left[\min\left(\rho_{i,t,k} A_{i,t},\ \operatorname{clip}\left(\rho_{i,t,k}, 1-\epsilon, 1+\epsilon\right) A_{i,t}\right)\right] + \beta\mathcal{L}_{\mathrm{KL}},
\label{eq:skillrubric_objective}
\end{equation*}
where given token $y_{i,t,k}$ with context
$\xi_{i,t,k}$, 
$
\rho_{i,t,k} = \frac{\pi_{\theta}(y_{i,t,k} \mid \xi_{i,t,k})}{\pi_{\theta_{\mathrm{old}}}(y_{i,t,k} \mid \xi_{i,t,k})}
$
is the importance ratio for token $k$ in turn $t$ of trajectory $i$, $\epsilon$ is the clipping threshold, and $\beta \mathcal{L}_{\mathrm{KL}}$ regularizes the policy toward the reference model.

\vspace{-2mm}
\subsection{Outcome-Grounded Alternating Skill Co-Evolution}
\label{sec:coevolution}
\vspace{-2mm}
The initial SkillBank $\mathcal{B}$ is distilled from teacher trajectories, but these demonstrations are static and cannot anticipate the novel strategies or failure modes that emerge as the policy learns. As the student policy improves, its own trajectories begin to reflect its current capabilities, revealing successful patterns absent from the teacher data and recurring pitfalls that the initial skills fail to address~\citep{xia2026skillrl,wu2026seed,wang2026agentopsd}. This creates a natural need for the SkillBank to evolve alongside the policy, ensuring that the guidance and rubrics remain aligned with the policy's ongoing experience rather than fixed to a frozen set of demonstrations.

Yet updating guidance and rubrics together poses a subtle but critical challenge: guidance shapes the trajectories the policy generates, while rubrics determine how those trajectories receive process credit. Changing both at once would make it impossible to attribute an observed improvement or regression to either factor, muddying the causal picture. To preserve interpretability and clean attribution, we therefore alternate their updates. For a skill $S^{(k)}=(d,g_k,\mathcal{C}_k)$, we follow the sequence
\begin{equation*}
(d,g_k,\mathcal{C}_k) \longrightarrow (d,g_{k+1},\mathcal{C}_k) \longrightarrow (d,g_{k+1},\mathcal{C}_{k+1}),
\label{eq:alternating_evolution}
\end{equation*}
while keeping the applicability $d$ and the outcome evaluator fixed. This ordering first isolates the effect of guidance on the trajectory distribution, and then adapts the rubric to the new distribution induced by the updated policy.

Concretely, each cycle follows a two-step alternating update procedure for the skills.
\vspace{-2mm}
\begin{itemize}[leftmargin=*,nosep]
    \item \textbf{Guidance update ($g_k \to g_{k+1}$).} We begin by comparing the fixed outcome verdict with the current rubric $\mathcal{C}_k$ on trajectories generated under $g_k$. 
    Their agreement or disagreement reveals specific deficiencies: joint failure may indicate insufficient guidance; rubric success with outcome failure suggests an over-permissive rubric; outcome success with rubric failure implies an overly strict rubric. 
    Guided by these signals, the synthesizer distills a candidate $\widetilde{g}$ only when recurring residual patterns support a guidance defect.
    To evaluate $\widetilde{g}$, we freeze the policy and compare its performance with $g_k$ under identical rollout conditions. We accept $\widetilde{g}$ as $g_{k+1}$ only if it improves overall outcomes without causing excessive regressions, where a regression means that $g_k$ succeeds but $\widetilde{g}$ fails; otherwise we retain $g_{k+1}=g_k$.   
    \item \textbf{Rubric update ($\mathcal{C}_k \to \mathcal{C}_{k+1}$).} After updating the guidance, we train the policy with $(g_{k+1},\mathcal{C}_k)$ and collect a fresh set of outcome-labeled trajectories. We then apply both the current rubric $\mathcal{C}_k$ and the candidate $\widetilde{\mathcal{C}}$ offline to these identical trajectories, isolating the effect of the rubric change. We accept $\widetilde{\mathcal{C}}$ as $\mathcal{C}_{k+1}$ if it aligns more closely with the fixed outcome verdict without achieving this agreement by uniformly relaxing its criteria; otherwise we retain $\mathcal{C}_{k+1}=\mathcal{C}_k$.
\end{itemize}
Repeating this alternating cycle progressively refines both guidance and supervision, while ensuring that each update remains grounded in final task outcomes. This co-evolutionary design allows the SkillBank to grow with the policy, turning the student's own experience into a source of continuous improvement rather than a one-time distillation.

\vspace{-3mm}
\section{Experiments}
\label{experiments}
\vspace{-3mm}











\subsection{Experimental Setup}
\label{sec:experimental_setup}
\vspace{-2mm}

\paragraph{Datasets.}
We evaluate our method on eight multimodal agent benchmarks. 
AgentVista~\citep{su2026agentvista}, BrowseComp-VL~\citep{geng2025webwatcher}, VDR-Bench~\citep{zeng2026vision}, and VisualToolBench~\citep{guo2025visualtoolbench} constitute the in-distribution setting.
We randomly sample 100 tasks from each benchmark for training and reserve disjoint tasks for evaluation, uniformly sampling 300 tasks from the remaining VDR-Bench pool. VisBrowse-Bench~\citep{zhang2026visbrowse}, MMSearch-Plus~\citep{tao2025mmsearch},
MM-SearchExam~\citep{zhang2026vsearcher}, and TIR-Bench~\citep{li2025tirbench} are used exclusively for out-of-distribution evaluation, with none of their tasks or trajectories included at any stage of training. Detailed dataset statistics and splits are provided in Appendix~\ref{app:dataset_details}.

\vspace{-2mm}
\paragraph{Implementation Details.}
We evaluate SkillRubric with Qwen3-VL-8B-Instruct~\citep{bai2025qwen3vl} and Qwen3.5-9B~\citep{qwen2026qwen35}. For each backbone, all RL methods share the same SFT initialization, constructed from success-filtered trajectories generated by GPT-5.5~\citep{openai2026gpt55}, Claude Opus 4.8~\citep{anthropic2026claudeopus48}, and Gemini 3.1 Pro~\citep{googledeepmind2026gemini31pro}, with three rollouts per teacher for each seed training task. 
We augment the 400 IID training tasks to 1k training instances through semantics-preserving instruction reformulation while retaining the original multimodal inputs, targets, and evaluation criteria.
The initial SkillBank is constructed from the 400 seed tasks, with cross-task skill clustering and merging performed independently within each benchmark before the resulting entries are pooled into a unified SkillBank.
This restriction prevents tasks with different objectives and evaluation semantics from being merged into overly generic guidance and rubrics.
We train for one epoch with 8 task instances per batch and sample 8 on-policy trajectories for each instance.
All agents operate in a unified tool environment comprising zoom, code interpreter, web search, image search, and webpage visit~\citep{jiang2026xskill}. 
SkillRubric retrieves the top-1 skill to maintain a unique correspondence between its actor-facing guidance, evaluator-facing rubric, and evolution target.
GPT-5-mini~\citep{openai2025gpt5} serves as the fixed training-time outcome evaluator and coarse-stage rubric matcher, while Gemini 3.5 Flash performs fine-grained multimodal verification.
We optimize the policy using a clipped GRPO objective with turn-level advantages and alternate guidance and rubric evolution every 10 optimization steps.
During evaluation, we generate one trajectory per test task using fixed decoding parameters under the same system prompt, tool environment, and tool-call limits, and append the top-1 guidance retrieved from the final SkillBank only for our skill-guided variant. DeepSeek-V4-Flash~\citep{xu2026deepseek} assigns a binary correctness label to each final prediction against the reference answers, with pass@1 computed as the mean correctness across tasks, whereas VisualToolBench reports the mean of its criterion-level judgments.
Full experimental settings and implementation details are provided in Appendices~\ref{app:tool_environment}--\ref{app:evaluation}.

\vspace{-3mm}
\paragraph{Baselines.}
We compare SkillRubric with four representative skill-based methods: XSkill~\citep{jiang2026xskill}, SkillRL~\citep{xia2026skillrl}, Skill-SD~\citep{wang2026skillsd}, and SKILL0~\citep{lu2026skill0}. Since these methods were not originally evaluated on our benchmark suite, we rerun their released implementations using the default hyperparameters reported in the corresponding papers. Detailed adaptations and baseline configurations are provided in Appendix~\ref{app:baseline_adaptation}.

\vspace{-3mm}
\subsection{Main Results}
\vspace{-2mm}

\begin{table*}[t]
\centering
\setlength{\tabcolsep}{4pt}
\renewcommand{\arraystretch}{1.15}
\caption{\textbf{Main results across eight multimodal agent benchmarks (\%).}
Each result is based on a single rollout per task.
All trained methods are initialized from the corresponding backbone-specific
SFT checkpoint and evaluated using the same tool suite and evaluation protocol.
Best and second-best results among trained methods under each backbone are
shown in \textbf{bold} and \underline{underline}, respectively.
Closed- and open-source reference models are excluded from ranking.
For our method, ``w/ skill'' and ``w/o skill'' denote inference with and
without skill guidance, respectively.}\vspace{-3mm}
\label{tab:main_results}
\resizebox{\textwidth}{!}{%
\begin{tabular}{l|cccc|cccc|c}
\toprule
\multirow{2}{*}{\textbf{Method}}
& \multicolumn{4}{c|}{\textbf{IID Benchmarks}}
& \multicolumn{4}{c|}{\textbf{OOD Benchmarks}}
& \multirow{2}{*}{\textbf{Avg.}} \\
\cmidrule(lr){2-5}\cmidrule(lr){6-9}
& AgentVista & BrowseComp-VL & VDR-Bench & VisualToolBench
& VisBrowse & MMSearch+ & MMSearchExam & TIR-Bench & \\
\midrule

\multicolumn{10}{l}{\emph{Closed-source model references}} \\

Gemini 3.1 Flash-Lite
& 26.61 & 43.81 & 20.00 & 31.01
& 33.14 & 33.76 & 26.15 & 27.00
& 30.19 \\


GPT-5 nano
& 12.84 & 35.79 & 5.67 & 14.95
& 24.85 & 5.14 & 24.73 & 17.50
& 17.68 \\

GPT-5 mini
& 16.51 & 45.15 & 11.67 & 27.54
& 36.09 & 14.79 & 34.98 & 24.50
& 26.40 \\


\midrule

\multicolumn{10}{l}{\emph{Open-source model references}} \\

Qwen3.5-35B-A3B
& 13.76 & 45.15 & 9.00 & 18.38
& 24.85 & 16.08 & 27.92 & 29.50
& 23.08 \\

Qwen3.5-122B-A10B
& 18.35 & 50.84 & 15.33 & 24.01
& 27.81 & 21.22 & 33.92 & 39.00
& 28.81 \\

Qwen3.6-35B-A3B
& 16.51 & 47.83 & 9.67 & 23.81
& 34.91 & 21.86 & 27.92 & 37.00
& 27.44 \\

\midrule

\multicolumn{10}{l}{\emph{Qwen3-VL-8B backbone}} \\

\rowcolor{gray!5}
Qwen3-VL-8B-Instruct
& 2.75 & 17.39 & 6.33 & 12.50
& 4.14 & 5.14 & 2.83 & 9.50
& 7.57 \\

SkillRL
& 7.34 & 23.08
& 10.67 & 15.00
& 14.20 & 10.29
& 9.89 & 14.50
& 13.12 \\

XSKILL
& 6.42 & 22.41 & 9.33 & 14.50
& 13.02 & 10.93
& 7.77 & 14.00
& 12.30 \\

Skill-SD
& 5.50 & 18.39 & 7.33 & 13.50
& 10.06 & 8.68 & 8.83 & 11.00
& 10.41 \\

SKILL0 (w/o skill)
& 6.42 & 22.07 & 8.67 & 13.00
& 11.83 & 7.72 & 5.30 & 11.50
& 10.81 \\

\rowcolor{blue!4}
\textbf{Ours (w/o skill)}
& \underline{8.26} & \underline{28.09}
& \underline{13.67} & \underline{16.00}
& \underline{17.16} & \underline{11.58}
& \underline{12.01} & \underline{16.50}
& \underline{15.41} \\

\rowcolor{blue!8}
\textbf{Ours (w/ skill)}
& \textbf{11.01} & \textbf{34.45}
& \textbf{16.67} & \textbf{17.50}
& \textbf{18.93} & \textbf{12.86}
& \textbf{14.13} & \textbf{17.50}
& \textbf{17.88} \\

\midrule

\multicolumn{10}{l}{\emph{Qwen3.5-9B backbone}} \\

\rowcolor{gray!5}
Qwen3.5-9B
& 11.93 & 37.12 & 7.67 & 17.50
& 24.85 & 11.25 & 20.85 & 19.50
& 18.83 \\

SkillRL
& 17.43 & 42.47
& 18.00 & 23.50
& 28.99 & 17.68
& 26.86 & 27.00
& 25.24 \\

XSKILL
& 16.51 & 40.47 & 16.33 & 22.00
& 27.22 & 18.33
& 27.21 & 26.50
& 24.32 \\

Skill-SD
& 15.60 & 38.46 & 16.67 & 21.50
& 26.63 & 15.76 & 24.73 & 24.00
& 22.92 \\

SKILL0 (w/o skill)
& 16.51 & 39.13 & 15.67 & 21.00
& 26.04 & 16.72 & 25.09 & 25.00
& 23.15 \\

\rowcolor{blue!4}
\textbf{Ours (w/o skill)}
& \underline{18.35} & \underline{45.48}
& \underline{21.00} & \underline{25.00}
& \underline{30.18} & \underline{20.26}
& \underline{28.98} & \underline{27.50}
& \underline{27.09} \\

\rowcolor{blue!8}
\textbf{Ours (w/ skill)}
& \textbf{19.27} & \textbf{46.82}
& \textbf{22.33} & \textbf{26.00}
& \textbf{31.95} & \textbf{20.90}
& \textbf{30.04} & \textbf{33.00}
& \textbf{28.79} \\

\bottomrule
\end{tabular}%
}\vspace{-5mm}
\end{table*}

As shown in \cref{tab:main_results}, SkillRubric achieves the best performance among trained methods across all eight benchmarks and both backbones, outperforming the strongest baseline, SkillRL, by 4.76 percentage points on Qwen3-VL-8B and 3.55 percentage points on Qwen3.5-9B on average.
These consistent improvements across IID and OOD benchmarks demonstrate the effectiveness and generalizability of jointly using actor-facing guidance and evaluator-facing rubrics during policy optimization. 
The gains cannot be attributed solely to additional guidance at inference, as our policies without skill guidance still outperform all competing training methods on both backbones and every benchmark. 
This result indicates that skill-conditioned optimization with turn-level rubric supervision enables the policy to internalize reusable planning and tool-use behaviors, while the further improvements obtained by restoring skill guidance demonstrate its complementary benefit. 
Notably, our Qwen3.5-9B policy reaches an average success rate of 28.79\%, nearly matching the much larger Qwen3.5-122B-A10B reference at 28.81\% and outperforming Qwen3.6-35B-A3B~\citep{qwen36_35b_a3b} at 27.44\%, as well as the closed-source GPT-5 nano and GPT-5 mini references at 17.68\% and 26.40\%. It also remains competitive with Gemini 3.1 Flash-Lite~\citep{google2026gemini31flashlite} (28.79\% vs.\ 30.19\%) and performs better on four of the eight benchmarks, demonstrating strong parameter efficiency and the ability of a compact open-source policy to compete with substantially larger open- and closed-source models.
To complement these quantitative results, we provide a \textbf{qualitative
comparison} of SkillRubric and the base model on a representative
VDR-Bench task in Appendix~\ref{app:case_vdr_sports},
with additional per-benchmark trajectory analyses
in Appendix~\ref{app:trajectory_analysis}.

\vspace{-4mm}
\subsection{Ablation Study}
\vspace{-3mm}

\begin{table*}[t]
\centering
\setlength{\tabcolsep}{7pt}
\renewcommand{\arraystretch}{0.9}
\caption{\textbf{Ablation study of the proposed components.}
All variants share the same backbone-specific SFT initialization and initial
SkillBank. IID, OOD, and Overall are unweighted averages over the corresponding
four, four, and all eight benchmarks, respectively.}\vspace{-3mm}
\label{tab:ablation}
\resizebox{\textwidth}{!}{%
\begin{tabular}{llcccccc}
\toprule
\multirow{2}{*}{\textbf{Backbone}}
& \multirow{2}{*}{\textbf{Variant}}
& \multicolumn{3}{c}{\textbf{Components}}
& \multicolumn{3}{c}{\textbf{Average Success Rate (\%)}} \\
\cmidrule(lr){3-5}\cmidrule(lr){6-8}
& & Outcome RL & Rubric Reward & Co-Evolution
& IID & OOD & Overall \\
\midrule

\multirow{4}{*}{Qwen3-VL-8B}
& SFT-only
& -- & -- & --
& 11.81 & 7.59 & 9.70 \\

& GRPO-only
& $\checkmark$ & -- & --
& 13.38 & 9.70 & 11.54 \\

& GRPO-Rubric
& $\checkmark$ & $\checkmark$ & --
& 15.46 & 11.70 & 13.58 \\

\rowcolor{blue!8}
& \textbf{Ours}
& $\checkmark$ & $\checkmark$ & $\checkmark$
& \textbf{19.91} & \textbf{15.86} & \textbf{17.88} \\

\midrule

\multirow{4}{*}{Qwen3.5-9B}
& SFT-only
& -- & -- & --
& 21.77 & 21.60 & 21.68 \\

& GRPO-only
& $\checkmark$ & -- & --
& 23.64 & 23.77 & 23.71 \\

& GRPO-Rubric
& $\checkmark$ & $\checkmark$ & --
& 25.60 & 25.37 & 25.49 \\

\rowcolor{blue!8}
& \textbf{Ours}
& $\checkmark$ & $\checkmark$ & $\checkmark$
& \textbf{28.61} & \textbf{28.97} & \textbf{28.79} \\

\bottomrule
\end{tabular}%
}\vspace{-2mm}
\end{table*}

\paragraph{Component contributions.}
The component ablation in \cref{tab:ablation} shows consistent and cumulative improvements from each proposed component across both backbones and data splits. Outcome-based GRPO establishes a stronger policy than SFT alone, while the additional gains from turn-level rubric rewards confirm that process supervision provides useful credit beyond the terminal outcome. Skill co-evolution contributes the largest marginal improvement, indicating that fixed guidance and evaluation criteria become increasingly suboptimal as the policy evolves. By adapting them to the policy's changing behaviors and remaining failures, the alternating updates maintain their effectiveness throughout training. These results validate the complementary roles of outcome optimization, rubric-based process supervision, and skill co-evolution.






\begin{table*}[t]
\centering
\setlength{\tabcolsep}{7pt}
\renewcommand{\arraystretch}{0.85}
\caption{\textbf{Ablation on the SkillBank evolution interval and components.}
\(H\) denotes the number of policy-optimization steps between successive
alternating updates in the full method.
\(H=\infty\) keeps the SkillBank fixed and corresponds to GRPO-Rubric.
Guidance evolution only and rubric evolution only update the indicated
component while keeping the other fixed, with \(H=10\).}
\vspace{-3mm}
\label{tab:ablation_evolution_interval}
\footnotesize
\begin{tabular}{llcccc}
\toprule
\textbf{Backbone}
& \textbf{Variant}
& \(\boldsymbol{H}\)
& \textbf{IID}
& \textbf{OOD}
& \textbf{Overall} \\
\midrule

\multirow{6}{*}{Qwen3-VL-8B}
& Ours & 5
& 16.34 & 12.79 & 14.57 \\
& Ours & 10
& 19.91 & 15.86 & 17.88 \\
& Ours & 20
& 18.35 & 13.98 & 16.17 \\
& GRPO-Rubric & \(\infty\)
& 15.46 & 11.70 & 13.58 \\
\cmidrule(lr){2-6}
& Guidance evolution only & 10
& 18.68 & 15.04 & 16.86 \\
& Rubric evolution only & 10
& 17.81 & 14.84 & 16.33 \\

\midrule

\multirow{6}{*}{Qwen3.5-9B}
& Ours & 5
& 26.43 & 26.19 & 26.31 \\
& Ours & 10
& 28.61 & 28.97 & 28.79 \\
& Ours & 20
& 27.21 & 26.72 & 26.97 \\
& GRPO-Rubric & \(\infty\)
& 25.60 & 25.37 & 25.49 \\
\cmidrule(lr){2-6}
& Guidance evolution only & 10
& 26.30 & 27.08 & 26.69 \\
& Rubric evolution only & 10
& 25.56 & 27.00 & 26.28 \\

\bottomrule
\end{tabular}
\vspace{-6mm}
\end{table*}

\vspace{-4mm}
\paragraph{Evolution interval and components.}
As shown in \cref{tab:ablation_evolution_interval},
\(H=10\) yields the best IID and OOD performance on both backbones
among the tested intervals.
This result suggests a trade-off between allowing policy adaptation
and maintaining timely SkillBank refinement.
With \(H=5\), the policy has fewer optimization steps to adapt
to revised guidance and rewards before the next evolution update,
so the resulting feedback may reflect transient adaptation
rather than persistent behavioral limitations.
Conversely, \(H=20\) may delay updates to guidance and rubrics
as new failure modes emerge.
All tested finite intervals outperform the fixed SkillBank,
supporting the benefit of continued evolution.
The component ablation further shows that evolving both components
outperforms either component alone.
Guidance shapes trajectory generation, whereas rubrics determine
how trajectories are evaluated during policy optimization.
Updating only one component may thus leave the other less aligned
with the policy's changing behavior, limiting the benefit of evolution.
These results support the complementary roles of guidance
and rubric adaptation.

\vspace{-4mm}
\subsection{SkillBank Evolution, Transferability, and Efficiency}
\vspace{-3mm}

\begin{figure*}[t]
    \centering
    \captionsetup[subfigure]{font=footnotesize,skip=2pt}


    \begin{subfigure}[t]{0.243\textwidth}
        \centering
        \includegraphics[width=\linewidth]
        {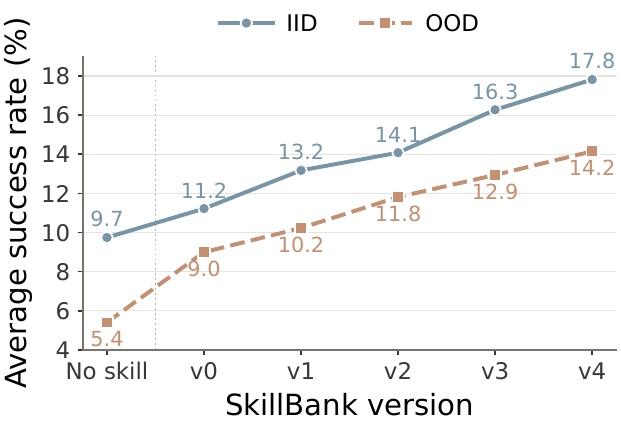}
        \caption{Qwen3-VL-8B-Instruct}\vspace{-1mm}
        \label{fig:skillbank_qwen3vl_8b}
        \vspace{-1mm}
    \end{subfigure}
    \hfill
    \begin{subfigure}[t]{0.243\textwidth}
        \centering
        \includegraphics[width=\linewidth]
        {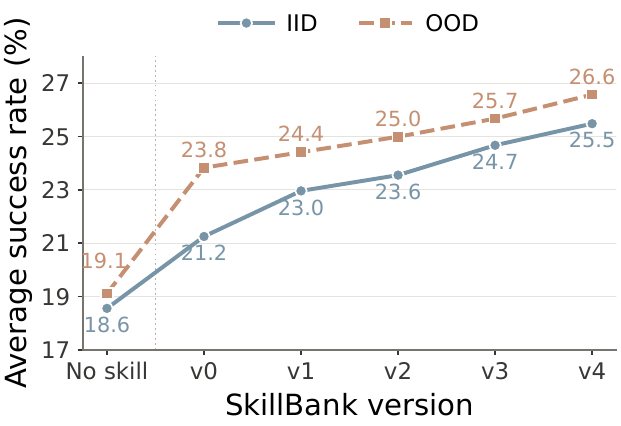}
        \caption{Qwen3.5-9B}\vspace{-1mm}
        \label{fig:skillbank_qwen35_9b}
        \vspace{-1mm}
    \end{subfigure}
    \hfill
    \begin{subfigure}[t]{0.243\textwidth}
        \centering
        \includegraphics[width=\linewidth]
        {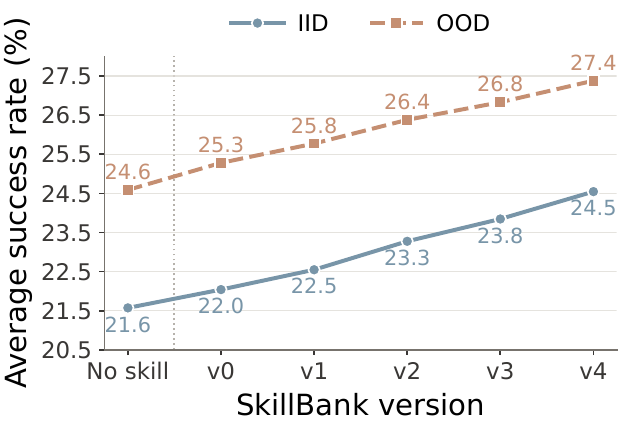}
        \caption{Qwen3.5-35B-A3B}\vspace{-1mm}
        \label{fig:skillbank_qwen35_35b}
        \vspace{-1mm}
    \end{subfigure}
    \hfill
    \begin{subfigure}[t]{0.243\textwidth}
        \centering
        \includegraphics[width=\linewidth]
        {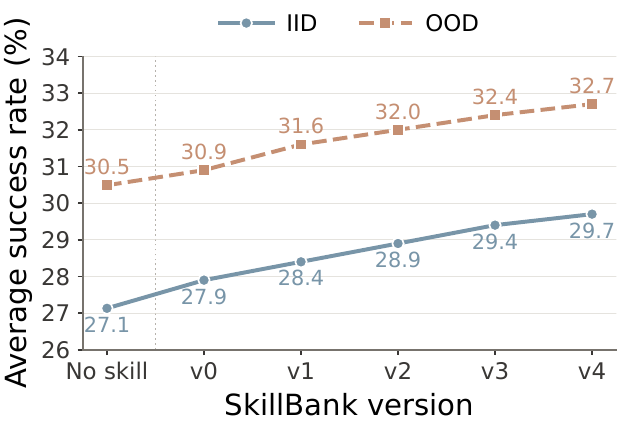}
        \caption{Qwen3.5-122B-A10B}\vspace{-1mm}
        \label{fig:skillbank_qwen35_122b}
        \vspace{-1mm}
    \end{subfigure}

    \par\vspace{2mm}


    \begin{subfigure}[t]{0.249\textwidth}
        \centering
        \includegraphics[width=\linewidth]
        {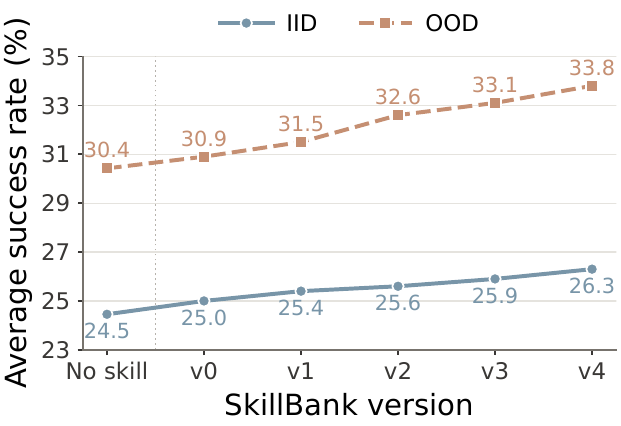}
        \caption{Qwen3.6-35B-A3B}\vspace{-2mm}
        \label{fig:skillbank_qwen36_35b}
        \vspace{-1mm}
    \end{subfigure}
    \hspace{0.025\textwidth}
    \begin{subfigure}[t]{0.249\textwidth}
        \centering
        \includegraphics[width=\linewidth]
        {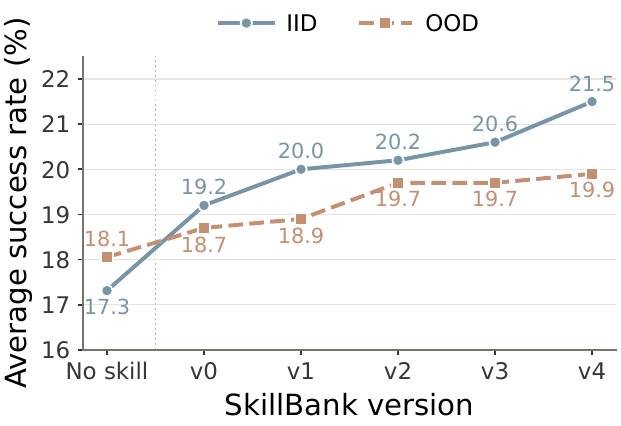}
        \caption{GPT-5 nano}\vspace{-2mm}
        \label{fig:skillbank_gpt5_nano}
        \vspace{-1mm}
    \end{subfigure}
    \hspace{0.025\textwidth}
    \begin{subfigure}[t]{0.249\textwidth}
        \centering
        \includegraphics[width=\linewidth]
        {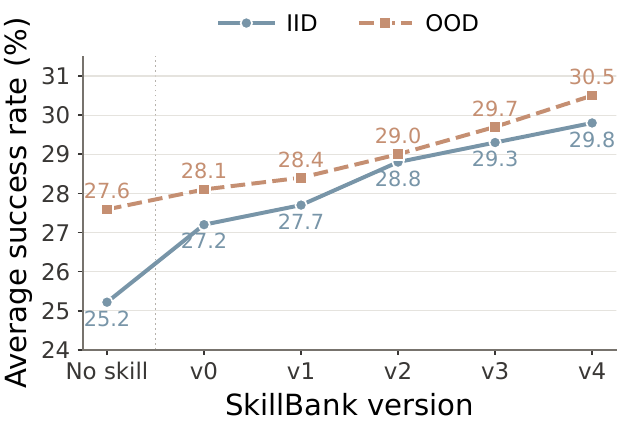}
        \caption{GPT-5 mini}\vspace{-2mm}
        \label{fig:skillbank_gpt5_mini}
        \vspace{-1mm}
    \end{subfigure}

    \caption{
        \textbf{Performance across SkillBank evolution.}
        Each panel reports the average success rates on IID and OOD benchmarks without skill guidance and with successive SkillBank versions \(v_0\)--\(v_4\). Qwen3-VL-8B and Qwen3.5-9B use the versions evolved with their respective policies, while all other models use those evolved with Qwen3.5-9B. Successive versions consistently improve performance across model families and task distributions.
    }
    \label{fig:skillbank_evolution}
    \vspace{-3mm}
\end{figure*}

\paragraph{SkillBank evolution and transferability.}

Figure~\ref{fig:skillbank_evolution} examines the effectiveness and transferability of successive SkillBank versions. Qwen3-VL-8B and Qwen3.5-9B are evaluated using the SkillBank versions evolved during their respective training processes. All other models use the versions evolved with Qwen3.5-9B, whose stronger base capability provides a higher-quality SkillBank sequence for transfer evaluation. This setting evaluates within-family transfer across different scales and architectures using Qwen3.5-35B-A3B and Qwen3.5-122B-A10B, as well as cross-family transfer to Qwen3.6-35B-A3B and the closed-source GPT models. The initial SkillBank consistently improves over inference without skill guidance, while successive versions yield further gains on both IID and OOD benchmarks across all evaluated models. These consistent trends show that co-evolution progressively improves the utility of skill guidance and that the resulting procedural knowledge transfers beyond the policy used for SkillBank evolution.

\begin{table*}[t]
\centering
\setlength{\tabcolsep}{4pt}
\renewcommand{\arraystretch}{0.95}
\caption{\textbf{Average interaction turns across eight multimodal agent
benchmarks.} Our models are conditioned on guidance retrieved from the final
SkillBank at inference, while the backbone models are evaluated without skill
guidance.}\vspace{-3mm}
\label{tab:average_turns}
\resizebox{\textwidth}{!}{%
\begin{tabular}{l|cccc|cccc|c}
\toprule
\multirow{2}{*}{\textbf{Model}}
& \multicolumn{4}{c|}{\textbf{IID Benchmarks}}
& \multicolumn{4}{c|}{\textbf{OOD Benchmarks}}
& \multirow{2}{*}{\textbf{Avg.}} \\
\cmidrule(lr){2-5}\cmidrule(lr){6-9}
& AgentVista
& BrowseComp-VL
& VDR-Bench
& VisualToolBench
& VisBrowse
& MMSearch+
& MMSearchExam
& TIR-Bench
& \\ 
\midrule

Qwen3-VL-8B-Instruct
& 6.7
& 4.6
& 3.9
& 8.4
& 5.8
& 5.0
& 4.6
& 7.9
& 5.9 \\

\rowcolor{gray!10}
Ours (Qwen3-VL-8B, w/ skill)
& 5.9
& 3.3
& 3.1
& 6.1
& 5.2
& 3.2
& 3.1
& 6.0
& \textbf{4.5} \\

Qwen3.5-9B
& 10.7
& 6.8
& 9.6
& 9.5
& 11.3
& 9.7
& 9.6
& 8.9
& 9.5 \\

\rowcolor{gray!10}
Ours (Qwen3.5-9B, w/ skill)
& 4.8
& 4.7
& 5.7
& 3.1
& 6.4
& 4.6
& 5.5
& 1.6
& \textbf{4.6} \\

\bottomrule
\end{tabular}%
}\vspace{-5mm}
\end{table*}

\vspace{-4mm}
\paragraph{Interaction efficiency.}
Table~\ref{tab:average_turns} evaluates the interaction efficiency of
SkillRubric. Our policies require fewer turns on every benchmark under both
backbones, reducing the overall average from 5.86 to 4.49 turns for
Qwen3-VL-8B and from 9.51 to 4.55 turns for Qwen3.5-9B. The reductions are
consistent across IID and OOD benchmarks, indicating that skill guidance
helps the policy identify effective action sequences while avoiding redundant
exploration and unproductive interactions. Together with the higher success
rates in \cref{tab:main_results}, these results show that SkillRubric improves
both task effectiveness and interaction efficiency.

\vspace{-4mm}
\section{Conclusion}
\vspace{-4mm}

Our work introduces \textbf{SkillRubric}, which pairs actor-facing guidance with evaluator-facing rubrics to turn reusable skills into both execution knowledge and evidence-grounded turn-level supervision. Its alternating co-evolution updates guidance and rubrics against fixed outcomes while isolating their effects. Across eight benchmarks and two backbones, SkillRubric consistently outperforms prior skill-based methods, transfers across model families, and reduces interaction turns, demonstrating that shared procedural knowledge can jointly improve agent behavior and process supervision.

\newpage
\bibliography{iclr2027_conference}

\newpage
\appendix

\begin{center}
    {\Large\bfseries Appendix}
\end{center}

\section{Dataset Details}
\label{app:dataset_details}

We evaluate SkillRubric on eight multimodal agent benchmarks under
in-distribution (IID) and out-of-distribution (OOD) settings.
Table~\ref{tab:dataset_statistics} summarizes the benchmark statistics and
data partitions. AgentVista, BrowseComp-VL, VDR-Bench, and VisualToolBench
constitute the IID setting. 
We sample 100 seed training tasks from each benchmark, yielding 400 seed tasks in total. Their teacher-generated trajectories are used for SFT initialization and dual-view SkillBank construction. For reinforcement learning, we augment these seed tasks into 1,000 task instances through semantics-preserving instruction reformulation while retaining the original multimodal inputs, targets, and evaluation criteria. 
The corresponding evaluation tasks are disjoint from the training split.

VisBrowse-Bench, MMSearch-Plus, MM-SearchExam, and TIR-Bench constitute the
OOD setting. None of their tasks or trajectories is used for SFT, SkillBank
construction, or reinforcement learning. We evaluate on the complete
VisBrowse-Bench, MMSearch-Plus, and MM-SearchExam datasets and sample 200 tasks
from TIR-Bench. Benchmark samples not selected for either training or
evaluation are not used elsewhere in our experiments. Unless otherwise
specified, all sampled partitions are constructed with random seed 42.

\textbf{AgentVista}~\citep{su2026agentvista} is a comprehensive benchmark for
evaluating generalist multimodal agents on realistic visual tasks. It contains
209 expert-curated tasks spanning 25 subdomains and seven broad categories,
including commerce, geography, entertainment, technology, society, academics,
and culture. Each task is grounded in one or more visual inputs and typically
requires long-horizon interaction with tools such as web search, image search,
webpage navigation, and code-based image processing. We randomly sample 100
tasks for training and use the remaining 109 tasks for IID evaluation.

\textbf{BrowseComp-VL}~\citep{geng2025webwatcher} extends BrowseComp-style
information-seeking tasks to the multimodal setting. It contains 399 tasks
across two difficulty levels: 199 Level-1 tasks requiring challenging
multi-hop retrieval and reasoning, and 200 Level-2 tasks in which key entities
or attributes are further obscured to prevent direct lookup. Solving these
tasks requires agents to combine visual perception with persistent web search,
evidence aggregation, and multi-step reasoning. We randomly sample 100 tasks
for training and use the remaining 299 tasks for IID evaluation.

\textbf{VDR-Bench}~\citep{zeng2026vision} evaluates multimodal deep-research
systems under realistic visual-retrieval conditions. It contains 2,000
expert-curated visual question-answering tasks designed to reduce shortcuts
arising from textual cues, model-internal knowledge, or near-exact image
matches. The tasks generally require iterative visual localization,
image-region inspection, textual retrieval, and cross-modal evidence
integration. We randomly sample 100 tasks for training and uniformly sample
300 disjoint tasks from the remaining pool for IID evaluation. The other
1,600 tasks are not used in our experiments.

\textbf{VisualToolBench}~\citep{guo2025visualtoolbench} evaluates whether multimodal models can actively manipulate and reason over visual content rather than treating images as static context. It comprises 1,204 open-ended tasks, including 603 single-turn and 601 multi-turn instances across five domains. These tasks involve operations such as cropping, enhancement, transformation, measurement, and programmatic image analysis, often combined with multi-step reasoning. We restrict our experiments to the single-turn subset, randomly sampling 100 tasks for training and a disjoint set of 200 tasks for IID evaluation. The remaining 904 tasks, including all multi-turn instances, are not used.

\textbf{VisBrowse-Bench}~\citep{zhang2026visbrowse} evaluates visual-native
browsing agents whose reasoning must remain grounded in visual evidence
throughout the search process. Its 169 human-verified tasks require agents to
retrieve, inspect, and cross-validate information from textual and visual
sources rather than relying only on textual search results or a one-time
recognition of the input image. We use all 169 tasks exclusively for OOD
evaluation.

\textbf{MMSearch-Plus}~\citep{tao2025mmsearch} contains 311 multimodal
browsing tasks spanning diverse domains, including geography, sports,
academia, film and television, technology, games, vlogs, and music. The
benchmark emphasizes fine-grained visual cue extraction, iterative
image--text retrieval, and source verification under noisy search results.
We use the complete benchmark exclusively for OOD evaluation.

\textbf{MM-SearchExam}~\citep{zhang2026vsearcher} evaluates long-horizon
multimodal web-search agents using 283 tasks that combine visual inputs with
complex, multi-hop information-seeking questions. Solving these tasks requires
extracting relevant visual clues, formulating successive search queries,
navigating webpages, and integrating evidence collected across multiple
sources. We use all 283 tasks exclusively for OOD evaluation.

\textbf{TIR-Bench}~\citep{li2025tirbench} evaluates agentic
thinking-with-images across 13 task categories that require active,
tool-assisted visual reasoning. Its tasks cover image enhancement, geometric
transformation, pixel-level analysis, spatial reasoning, visual puzzles, and
iterative image manipulation. We sample 200 tasks for OOD evaluation, with no
TIR-Bench data used during SFT, SkillBank construction, or reinforcement
learning.

\begin{table*}[!t]
\centering
\small
\caption{\textbf{Dataset statistics and train/evaluation splits.}
Available denotes the size of the complete benchmark, while Train and Eval
report the subsets used in our experiments. Unselected samples are not used
elsewhere, and all OOD evaluation tasks are excluded from training.}
\label{tab:dataset_statistics}
\setlength{\tabcolsep}{6pt}
\renewcommand{\arraystretch}{1.08}
\begin{tabular}{llcrrr}
\toprule
\textbf{Dataset}
& \textbf{Domain}
& \textbf{Setting}
& \textbf{Available}
& \textbf{Train}
& \textbf{Eval} \\
\midrule

AgentVista
& Comprehensive Multimodal Tool Use
& IID
& 209
& 100
& 109 \\

BrowseComp-VL
& Multimodal Browsing
& IID
& 399
& 100
& 299 \\

VDR-Bench
& Multimodal Deep Research
& IID
& 2,000
& 100
& 300 \\

VisualToolBench
& Tool-Assisted Visual Reasoning
& IID
& 1,204
& 100
& 200 \\

\midrule

VisBrowse-Bench
& Visual-Native Browsing
& OOD
& 169
& --
& 169 \\

MMSearch-Plus
& Multimodal Search
& OOD
& 311
& --
& 311 \\

MM-SearchExam
& Multimodal Web Search
& OOD
& 283
& --
& 283 \\

TIR-Bench
& Tool-Assisted Visual Reasoning
& OOD
& 1,215
& --
& 200 \\

\bottomrule
\end{tabular}
\end{table*}

\section{Implementation Details}
\label{app:implementation_details}

\subsection{Tool Environment}
\label{app:tool_environment}

\begin{table*}[t]
\centering
\small
\caption{\textbf{Tool definitions and execution settings.}
All compared methods use the same tools and per-trajectory call limits.}
\label{tab:tool_definitions}
\setlength{\tabcolsep}{5pt}
\renewcommand{\arraystretch}{1.08}
\begin{tabularx}{\linewidth}{
    @{}
    >{\raggedright\arraybackslash}p{0.14\linewidth}
    >{\raggedright\arraybackslash}X
    >{\raggedright\arraybackslash}p{0.20\linewidth}
    >{\raggedright\arraybackslash}p{0.14\linewidth}
    @{}
}
\toprule
\textbf{Tool}
& \textbf{Function}
& \textbf{Backend}
& \textbf{Call limit} \\
\midrule

Zoom
& Crops or enlarges selected image regions for fine-grained visual inspection.
& Local image processor
& No separate limit \\

Code interpreter
& Executes Python code for image processing, numerical computation, and data analysis.
& Isolated code sandbox
& No separate limit \\

Web search
& Retrieves relevant webpages and textual search results for a text query.
& SerpAPI
& 7 per trajectory \\

Image search
& Retrieves visually related images and their associated webpages and sources.
& SerpAPI; ImgBB for temporary hosting
& 5 per trajectory \\

Visit
& Opens a specified webpage and extracts its textual and visual content.
& Jina and Firecrawl
& No separate limit \\

\bottomrule
\end{tabularx}
\end{table*}

All policies interact with the same unified tool environment through
schema-validated function calls. The environment provides five tools:
\emph{zoom}, \emph{code interpreter}, \emph{web search}, \emph{image search},
and \emph{visit}. Table~\ref{tab:tool_definitions} summarizes their
functionality, execution backends, and call limits. Each rollout contains at
most 20 assistant turns. The code interpreter supports common image-processing,
numerical, and data-analysis libraries, including PIL, OpenCV, NumPy, SciPy,
Matplotlib, and Pandas, and can return generated images to the policy as
subsequent visual observations. ImgBB provides temporary image hosting when an
external service requires an image URL.

To control context growth during multi-turn interaction, textual observations
returned by web search and webpage visits are summarized before being inserted
into the policy context when they exceed 2,000 characters. We use Gemini 3.1
Flash-Lite with a maximum summary length of 4,096 tokens. This compression
affects only the policy context: the original uncompressed observations and
visual evidence remain available to the multimodal verifier for evidence
checking.

\subsection{Training and Optimization}
\label{app:training_optimization}

\paragraph{Training data and SFT initialization.}
We use Qwen3-VL-8B-Instruct~\citep{bai2025qwen3vl} and
Qwen3.5-9B~\citep{qwen2026qwen35} as the student backbones. For each of the
400 IID training tasks, GPT-5.5~\citep{openai2026gpt55}, Claude Opus
4.8~\citep{anthropic2026claudeopus48}, and Gemini 3.1
Pro~\citep{googledeepmind2026gemini31pro} each generate three trajectories.
All successful and failed trajectories are used to construct the initial
SkillBank $\mathcal{B}^{(0)}$, while approximately 2,000 success-filtered
trajectories are retained for SFT.

Both backbones are trained for one epoch using AdamW with a learning rate of
$5\times10^{-6}$, weight decay $0.01$, a cosine learning-rate schedule, and a
warmup ratio of $0.1$. We use an effective SFT batch size of 16 and a maximum
sequence length of 32,768 tokens. All non-vision parameters are updated, while
the visual encoder is frozen. Training is performed in bfloat16. All RL
methods using the same backbone are initialized from the corresponding SFT
checkpoint.

\paragraph{Skill construction and retrieval.}
The frozen skill synthesizer \(\mathcal{M}_{\mathrm{skill}}\) is instantiated
with GPT-5.5 for both initial SkillBank construction and subsequent
co-evolution proposals. We use a frozen Qwen3-Embedding-4B model as
\(f_{\mathrm{emb}}\). Skill applicability descriptions are embedded and
cached in advance, and all embeddings are \(\ell_2\)-normalized before
cosine-similarity retrieval. Given a task, we retrieve the top-1 skill from
the current SkillBank \(\mathcal{B}^{(v)}\), establishing a unique
correspondence among its applicability description, actor-facing guidance,
evaluator-facing rubric, and co-evolution target.

\paragraph{Policy prompting.}
All policies use the base system prompt in 
\cref{prompt:base_agent_system} during both training and evaluation. For a
skill-guided rollout, the retrieved actor-facing guidance \(g_j\) is
serialized into structured text and appended directly to the system message
under the \texttt{\# Skill Guidance} header, following the template in
\cref{prompt:skill_injection}. A final instruction emphasizes that the
guidance is non-prescriptive and may be replaced by any method that reaches
the same evidence state. The user message contains the original task question
and its input images. Neither the applicability description \(d_j\), the
paired evaluator-facing rubric \(\mathcal{C}_j\), nor verifier feedback is
exposed to the policy. Skill-free variants use the base system prompt without
the guidance suffix.



\refstepcounter{srprompt}
\label{prompt:base_agent_system}
\begin{skillrubricprompt}
{Prompt~\thesrprompt: Shared policy system prompt for training and evaluation}
You are a visual reasoning agent. Answer the user's question using the
provided images and the available tools.

Use tools when needed to inspect visual details, perform calculations,
or gather relevant evidence. You may call at most one tool in each
assistant turn. After a tool returns, inspect its result before deciding
on the next action.

Base your answer on the available evidence. When you have sufficient
evidence, provide the entire final answer inside exactly one complete
<answer>...</answer> block.
\end{skillrubricprompt}

\refstepcounter{srprompt}
\label{prompt:skill_injection}
\begin{skillrubricprompt}
{Prompt~\thesrprompt: Actor-Facing Skill Guidance Suffix}
# Skill Guidance

{formatted_actor_facing_guidance}

These are suggestions, not mandates. Use another method if it reaches the
same evidence state.
\end{skillrubricprompt}

\paragraph{Policy rollouts.}
We optimize each backbone for one epoch over the 1k training tasks. Each
rollout batch contains eight tasks, and \(G=8\) trajectories are sampled for
each task. The eight trajectories generated for one task constitute a GRPO
group. Thus, each rollout batch contains eight groups and 64 trajectories in
total. During training, trajectories are sampled with temperature \(0.7\) and
top-\(p=0.9\). Each trajectory contains at most 20 assistant turns, with
maximum context and response lengths of 32,768 tokens.

\paragraph{Policy objective.}
We optimize the policy using clipped GRPO with a learning rate of
$3\times10^{-6}$, clipping parameter $\epsilon=0.2$, and KL coefficient
$\beta_{\mathrm{KL}}=0.1$. Unless otherwise specified, rubric criteria use
equal weights $w_c=1$. We set the incremental-progress scaling factor to
$\omega=1$ and clip the process advantage to $[-\delta,\delta]$ with
$\delta=2$.
All policy-generated tokens within assistant turn $t$ share $A_{i,t}$, whereas
screenshots, tool outputs, and other environment-generated content are
excluded from the policy loss.

\subsection{Outcome and Rubric Evaluation}
\label{app:rubric_evaluation}



\paragraph{Outcome evaluation.}
GPT-5 mini serves as the fixed outcome evaluator during policy training and SkillBank co-evolution. For benchmarks with verifiable reference answers, it assigns a binary correctness score $r_i^{\mathrm{out}}\in\{0,1\}$ by comparing the final prediction with the acceptable reference answers. For VisualToolBench, it judges each benchmark-provided criterion independently and defines
$r_i^{\mathrm{out}}$ as the mean of these binary judgments.
Trajectories without a valid final answer receive zero outcome reward.
The training-time outcome evaluator is independent of the retrieved SkillBank rubric $\mathcal{C}_j$ and remains fixed throughout training and co-evolution.
Final benchmark evaluation uses a separate DeepSeek-V4-Flash judge, as detailed in Appendix~\ref{app:evaluation}.

\paragraph{Multimodal rubric verification.}
We instantiate the coarse-stage selector with GPT-5-mini and the fine-grained
multimodal verifier with Gemini 3.5 Flash~\citep{google2026gemini35flash}. At the coarse stage, GPT-5-mini
summarizes each assistant turn using only the evidence available at that point,
determines criterion applicability, and identifies candidate turns where the
evidence state may have changed. At the fine stage, Gemini 3.5 Flash examines
the selected turns, their raw responses and observations, and the relevant
preceding evidence. For each applicable criterion \(c\), it produces a binary
satisfaction score \(s_{i,c}\in\{0,1\}\) and non-negative responsibility scores
\(\alpha_{i,c,t}\) over the selected turns. Criteria judged not applicable or
unverifiable are excluded from process-advantage computation.

To estimate incremental progress, the fine verifier classifies the evidence
state after each selected turn as \emph{missing}, \emph{hypothesis},
\emph{supported}, \emph{verified}, or \emph{contradicted}. We map
\emph{missing} and \emph{contradicted} to \(0\), and the remaining states to
\(0.25\), \(0.6\), and \(1\), respectively. We then compute
\(\Delta\phi_{i,c,t}\) from the change in the mapped evidence state introduced
by assistant turn \(t\). Positive, negative, and zero values indicate progress,
regression, and no verified change, respectively. The complete prompt templates
for the two stages are provided in
\cref{prompt:stage1_selector,prompt:stage2_verifier}.

\refstepcounter{srprompt}
\label{prompt:stage1_selector}
\begin{skillrubricprompt}
{Prompt~\thesrprompt: Stage 1 coarse selector}
You are Stage 1 of a two-stage rubric evaluator: a CAUSAL SEMANTIC INDEXER and EVIDENCE LOCATOR.

You receive the complete model-visible trajectory, segmented by turn. Long configured tool results may already have been condensed by the rollout summarizer; every other observation is the complete text that entered the policy context. Images may be attached and are identified by the image manifest.

Your job is to:
1. Produce a concise semantic card for EVERY assistant turn.
2. Conservatively determine checkpoint applicability.
3. Locate candidate turns where the checkpoint's evidence state may have changed.

You MUST NOT decide pass/fail, assign reward, or claim that a hypothesis was correct merely because a later turn confirmed it. Stage 2 alone judges and assigns credit.

## Question
{question}

## Skill and evidence guidance
{skill}

## Rubric checkpoints
{rubric}

## Complete segmented trajectory
{trajectory_steps}

## Image manifest
{image_manifest}

## Deterministic high-recall turn hints
{forced_turns}

## Causal / hindsight rules
- Summarize turn t using only information available by the END of turn t.
- Later evidence may resolve an earlier hypothesis, but must never be used to rewrite the earlier hypothesis as supported or verified.
- Distinguish claim status: "hypothesis", "supported", "verified", "contradicted", or "unknown".
- Every textual evidence reference must identify source_turn and include a short verbatim_quote from that turn.
- Every visual evidence reference must use an image_id from the manifest. Its available_at_turn must be <= the summarized/candidate turn.
- A tool executing successfully does not imply that its result is relevant or correct.
- Recall is more important than precision for candidates; uncertain applicability must not be filtered out.

Candidate change_type must be one of: "acquire", "verify", "regress", "contradict", "missed_opportunity", "finalize".
Only reference real turn ids. Keep summaries concise.

Return ONLY valid JSON following the illustrative schema below:
{
  "turn_summaries": [
    {
      "turn": 0,
      "action_summary": "<what the assistant attempted>",
      "belief_updates": [
        {
          "claim": "<claim at this turn>",
          "status": "hypothesis",
          "first_supported_turn": null,
          "first_verified_turn": null,
          "text_evidence_refs": [],
          "visual_evidence_refs": []
        }
      ],
      "unresolved": ["<what remains unknown>"]
    }
  ],
  "rubrics": {
    "C1": {
      "applicability": "active",
      "candidates": [
        {
          "turn": 4,
          "change_type": "acquire",
          "confidence": 0.82,
          "text_evidence_refs": [
            {
              "source_turn": 4,
              "verbatim_quote": "<exact quote>"
            }
          ],
          "visual_evidence_refs": ["turn_4_observation_1"]
        }
      ]
    }
  }
}
\end{skillrubricprompt}

\refstepcounter{srprompt}
\label{prompt:stage2_verifier}
\begin{skillrubricprompt}
{Prompt~\thesrprompt: Stage 2 multimodal verifier}
You are Stage 2 of a two-stage rubric evaluator: the CAUSAL VERIFIER and CREDIT ATTRIBUTOR.

Stage 1 produced semantic turn cards and candidate evidence turns. You now read the selected RAW model-visible evidence windows and any attached images. You are the ONLY stage allowed to decide checkpoint pass/fail and assign process credit.

## Question
{question}

## Skill
{skill}

## Rubric checkpoints
{rubric}

## Stage 1 applicability
{applicability}

## Stage 1 semantic trace
{semantic_trace}

## Candidate turn ids (the ONLY turns you may attribute to)
{candidate_turns}

## Raw evidence windows
{evidence_windows}

## Image manifest
{image_manifest}

## Final agent response
{response}

## Environment outcome
{outcome}

## Evidence-state scale
For each checkpoint, state changes use ONLY these ordered states:
- "missing" = no relevant supported state
- "hypothesis" = an unverified possibility was proposed
- "supported" = relevant evidence provides meaningful support
- "verified" = sufficiently reliable evidence verifies the state
- "contradicted" = prior support was invalidated or contradicted

## Strict causal rules
1. Judge turn t only from evidence available by the END of turn t. Later evidence cannot retroactively reward an earlier guess.
2. A positive state transition must cite raw textual evidence or an attached image that was available by event_turn.
3. text_evidence_refs require source_turn <= event_turn and a verbatim_quote from that source turn.
4. visual_evidence_refs require image_id.available_at_turn <= event_turn. If visual evidence is required but no relevant image is attached, use "unverifiable" rather than trusting the agent's visual claim.
5. Judge method-agnostically; do not reward a tool name, tool count, or operation order.
6. Both state_events and attribution must reference only candidate_turns. The role restrictions conditioned on the final verdict apply only to the attribution list: use "supporting" for pass, and "violation" or "missed_opportunity" for fail. State events independently record all verified progress and regression, including positive progress in a trajectory whose final checkpoint verdict is fail. Do not suppress a valid state transition because of the final verdict.
7. Final verdict is BINARY: "pass" or "fail". "not_applicable" and "unverifiable" are excluded from rubric reward. Do not output partial scores.
8. State events describe genuine evidence-state changes only. Do not emit an event for a neutral/redundant turn.

Return ONLY valid JSON following the illustrative schema below:
{
  "C1": {
    "verdict": "pass",
    "confidence": 0.9,
    "state_events": [
      {
        "turn": 4,
        "before": "missing",
        "after": "supported",
        "role": "supporting",
        "confidence": 0.86,
        "text_evidence_refs": [
          {
            "source_turn": 4,
            "verbatim_quote": "<exact quote>"
          }
        ],
        "visual_evidence_refs": []
      },
      {
        "turn": 7,
        "before": "supported",
        "after": "verified",
        "role": "supporting",
        "confidence": 0.91,
        "text_evidence_refs": [
          {
            "source_turn": 7,
            "verbatim_quote": "<exact quote>"
          }
        ],
        "visual_evidence_refs": []
      }
    ],
    "attribution": [
      {
        "turn": 4,
        "weight": 0.4,
        "role": "supporting"
      },
      {
        "turn": 7,
        "weight": 0.6,
        "role": "supporting"
      }
    ]
  }
}
\end{skillrubricprompt}


\paragraph{Degenerate rubric signals.}
If a criterion has zero satisfaction variance within a rollout group,
its normalized satisfaction term is set to zero, while valid
incremental-progress signals are retained. If a failed criterion has
a negative normalized satisfaction signal but receives zero total
responsibility, this signal is assigned to the final eligible decision
turn in its candidate set. A rollout group is removed only when its
outcome advantages and all turn-level process advantages are zero.

\subsection{SkillBank Co-Evolution}
\label{app:coevolution_details}

Guidance and rubric updates alternate every \(H=10\) policy optimization
steps. GPT-5.5 analyzes repeated high-confidence failure modes and
outcome--rubric residuals using the diagnostic prompt in
\cref{prompt:coevolution_coach}. It proposes a guidance patch only when the
residuals reveal a genuine guidance defect, whereas rubric revisions are
restricted to diagnosed ambiguity, coverage, applicability, verifiability,
weight-calibration, or judge-instability issues. The active rubric remains
fixed during proposal generation, preventing a candidate rubric from
validating its own guidance proposal. The applicability description \(d_j\)
and the outcome evaluator remain fixed throughout co-evolution.

\refstepcounter{srprompt}
\label{prompt:coevolution_coach}
\begin{skillrubricprompt}
{Prompt~\thesrprompt: Skill and rubric diagnosis coach}
You are the Skill/Rubric diagnosis coach in a policy/Skill co-evolution
system.

The ACTIVE rubric is immutable while this proposal is produced. A Candidate
Rubric may be admitted later only after offline re-judging and an
outcome-alignment gate; it never validates its own Skill candidate.

Revise Skill guidance only when the diagnosis shows an actual guidance
defect. For a pure rubric defect, set skill_patch to null instead of changing
guidance unnecessarily. Never change the rubric merely to make the policy or
a new Skill easier to pass.

Do not add task answers, labels, image-specific facts, mandatory tool
sequences, or method-specific requirements. Preserve generality and method
independence.

Allowed diagnosis.type values:
- skill_gap
- policy_execution_failure
- insufficient_evidence
- rubric_ambiguity
- rubric_coverage_gap
- rubric_applicability_gap
- rubric_verifiability_gap
- rubric_weight_miscalibration
- judge_instability

A rubric_proposal is allowed only for a rubric_* diagnosis or
judge_instability; otherwise it MUST be null.

Allowed rubric_proposal.operation values are:
- clarify
- add_checkpoint
- split_checkpoint
- merge_checkpoints
- applicability_fix
- verifiability_fix
- weight_calibration

Use applicability_fix only for rubric_applicability_gap,
weight_calibration only for rubric_weight_miscalibration, and
add_checkpoint only for rubric_coverage_gap.

If proposed, candidate_criteria MUST be the complete replacement criteria
object, not a partial diff. Preserve every source criteria top-level field
and set method_agnostic=true on every candidate checkpoint. For
clarify/applicability_fix/verifiability_fix/weight_calibration, preserve all
checkpoint IDs.

Return one JSON object only following one of the two illustrative schemas
below. Replace all example values with judgments based on the input.

For a guidance defect:
{
  "diagnosis": {
    "type": "skill_gap",
    "confidence": 0.0,
    "evidence_summary": "..."
  },
  "skill_patch": {
    "guidance": {
      "overview": "...",
      "stages": [...],
      "pitfall_warnings": [...]
    },
    "rationale": "..."
  },
  "rubric_proposal": null
}

For a rubric defect:
{
  "diagnosis": {
    "type": "rubric_ambiguity",
    "confidence": 0.0,
    "evidence_summary": "..."
  },
  "skill_patch": null,
  "rubric_proposal": {
    "diagnosis_type": "rubric_ambiguity",
    "confidence": 0.0,
    "operation": "clarify",
    "target_checkpoint_ids": ["C1"],
    "candidate_criteria": {
      "checkpoints": [...]
    },
    "rationale": "...",
    "evidence_summary": "...",
    "risks": []
  }
}

Current Skill:
{skill}

Repeated high-confidence failure modes:
{failures}
\end{skillrubricprompt}

\paragraph{Guidance validation.}
A guidance candidate is compared with the current guidance through matched
rollouts under a frozen policy. Each validation uses 8 tasks, with one rollout
generated under the current guidance and one under the candidate guidance for
each task using the same decoding configuration. At least 4 valid rollout
pairs are required. The candidate is accepted only when it improves the mean
outcome and does not produce a paired regression greater than \(0.05\).

\paragraph{Rubric validation.}
After guidance validation, the policy is trained for another \(H\)
optimization steps using the selected guidance \(g^{(v+1)}\) and the current
rubric \(\mathcal{C}^{(v)}\). This delay allows the trajectory distribution to
adapt to the updated guidance before the rubric is evaluated. We then apply
\(\mathcal{C}^{(v)}\) and the candidate rubric
\(\widetilde{\mathcal{C}}\) offline to the same outcome-labeled trajectories,
ensuring that their difference cannot be attributed to additional policy
updates or environment variation. Each comparison uses 128 cached trajectories.

Let \(a(\mathcal{C})\) denote the alignment between the weighted score
produced by rubric \(\mathcal{C}\) and the fixed outcome verdict, and let
\(p(\mathcal{C})\) denote its pass rate over the calibration trajectories. We
estimate the alignment gain
\[
\Delta a
=
a(\widetilde{\mathcal{C}})
-
a(\mathcal{C}^{(v)})
\]
using 1,000 paired bootstrap samples. The candidate passes the alignment test
when the one-sided \(90\%\) lower confidence bound satisfies
\[
\Delta a > 0,
\qquad
\operatorname{LCB}_{0.9}(\Delta a) \geq -0.02.
\]
The candidate must show a positive estimated alignment gain, while the lower confidence bound may extend to $-0.02$ to accommodate finite-sample uncertainty.

We additionally measure the pass-rate change
$
\Delta p
=
p(\widetilde{\mathcal{C}})
-
p(\mathcal{C}^{(v)})$
and require \(\Delta p\leq 0.10\). 
This constraint limits pass-rate inflation due to relaxed criteria.
If both conditions are satisfied, we set
\(\mathcal{C}^{(v+1)}=\widetilde{\mathcal{C}}\). Otherwise, we retain
\(\mathcal{C}^{(v+1)}=\mathcal{C}^{(v)}\). In either case, the previously
validated guidance \(g^{(v+1)}\) remains unchanged, and the resulting
SkillBank snapshot contains the updated skill
$
\bigl(d_j,g^{(v+1)},\mathcal{C}^{(v+1)}\bigr)$.

\subsection{Evaluation Details}
\label{app:evaluation}
\paragraph{Evaluation protocol.}
We evaluate on the test splits summarized in Table~\ref{tab:dataset_statistics}, using all test tasks of each benchmark. Policies are served with SGLang behind an OpenAI-compatible API and queried with a single rollout per task under near-greedy decoding. 
All models share the same base system prompt (\cref{prompt:base_agent_system}), tool environment, and tool-call limits as in training.
A trajectory must state its final answer inside \texttt{<answer>...</answer>} tags; trajectories without a valid final answer, including aborted and truncated ones, are scored as incorrect. Table~\ref{tab:eval_hyperparameters} lists the full configuration.

\paragraph{Judging.}

Final correctness is judged by DeepSeek-V4-Flash. Given the question, the reference answer, and the prediction, the judge returns a JSON verdict \texttt{\{"score", "explanation"\}}.
We use two judging templates: one for tasks with verifiable answers and one for criterion-based evaluation on VTB-Bench. 
For a single reference answer, the judge accepts semantic equivalence, including surface variation in capitalization, punctuation, and synonyms. 
For multiple-choice questions, the prediction must identify the correct option, with no numerical tolerance between options. A $\pm 5\%$ numerical tolerance applies only to free-form numerical answers.
For benchmarks with a set of acceptable answers, matching any reference is correct. For VTB-Bench, each rubric criterion is judged independently and the sample score is the average over its criteria; all other benchmarks report accuracy. 
The two judging templates are provided in \cref{prompt:judge_verifiable,prompt:judge_vtb_rubric}.

\paragraph{Skill injection during evaluation.}
During evaluation, we apply the same skill-retrieval procedure used in
training. For each evaluation task with description $q$, the fixed embedding
model $f_{\mathrm{emb}}$ retrieves the top-1 skill from the final SkillBank
$\mathcal{B}^{(V)}$ according to the cosine similarity between
$f_{\mathrm{emb}}(q)$ and each skill applicability embedding
$f_{\mathrm{emb}}(d_j)$. The retrieved actor-facing guidance $g_j$ is appended
to the policy prompt in the same format used for RL rollouts, while its paired
evaluator-facing rubric $\mathcal{C}_j$ remains hidden from the policy.

\begin{table}[t]
\centering
\small
\caption{Final evaluation configuration. The inference and judging settings are shared by all compared models; skill settings apply only to skill-conditioned models.}
\label{tab:eval_hyperparameters}
\setlength{\tabcolsep}{4pt}
\renewcommand{\arraystretch}{1.05}
\begin{tabularx}{\linewidth}{
    @{}
    >{\raggedright\arraybackslash}p{0.14\linewidth}
    >{\raggedright\arraybackslash}p{0.40\linewidth}
    >{\raggedright\arraybackslash}X
    @{}
}
\toprule
Group & Setting & Value \\
\midrule
\multirow{9}{*}{Inference}
& Serving backend & SGLang (OpenAI-compatible) \\
& Sampling temperature & 0.2 \\
& Top-$p$ & 1.0 \\
& Rollouts per task & 1 \\
& Maximum interaction turns & 20 \\
& Maximum context length & 131,072 tokens \\
& Maximum generated tokens & 32,768 \\
& Maximum images per task & 100 \\
& Web / image search call limits & 7 / 5 per trajectory \\
\midrule
\multirow{4}{*}{Judging}
& Judge model & DeepSeek-V4-Flash \\
& Judge output format & JSON \texttt{\{score, explanation\}} \\
& Judge \texttt{max\_tokens} & 256 (512 for rubric judging) \\
& Judge temperature & 0 \\
\bottomrule
\end{tabularx}
\end{table}


\paragraph{Evaluation prompts.}
Training and evaluation use the same base system prompt in
\cref{prompt:base_agent_system}. For skill-conditioned SkillRubric
inference, the retrieved actor-facing guidance is appended using
the template in \cref{prompt:skill_injection}. The two outcome-judging
templates are provided in
\cref{prompt:judge_verifiable,prompt:judge_vtb_rubric}.






\refstepcounter{srprompt}
\label{prompt:judge_verifiable}
\begin{skillrubricprompt}
{Prompt~\thesrprompt: Judge template for tasks with verifiable answers}
You are an impartial judge. You will be given a question, a set of reference answers, and a model's prediction.

Evaluation rules:
- The prediction is correct if it matches ANY ONE of the reference answers semantically.
- The prediction can be correct even if it includes extra information, as long as it contains a reference answer or clearly conveys the same meaning.
- Accept minor variations in capitalization, punctuation, and synonyms.
- Answers written in a different language than the question are acceptable if they are semantically equivalent to the reference answer.
- For multiple-choice questions, the prediction must identify the correct option, either by its label or by semantically equivalent option content. Numerical tolerance does not apply to choosing among options.
- For free-form numerical answers, accept small numerical differences within 5% of the reference value.
- More specific answers (e.g., "top three" vs "first place") should be judged accordingly.
- Judge only correctness, not the reasoning process.

Output your evaluation in the following JSON format:
{
  "score": 1 or 0,
  "explanation": "a short explanation of why the prediction is correct or incorrect"
}

Be strict. Assign score 1 only if the prediction is correct.

Question:
{question}

Reference Answers:
{answer}

Prediction:
{prediction}

Evaluate now. Output only valid JSON.
\end{skillrubricprompt}

\refstepcounter{srprompt}
\label{prompt:judge_vtb_rubric}
\begin{skillrubricprompt}
{Prompt~\thesrprompt: Judge template for VTB rubric criteria}
You are an impartial judge. You will be given a question, a rubric criterion, and a model's prediction.

Evaluation rules:
- The criterion is considered satisfied if the prediction meets it semantically.
- The prediction can include extra information, as long as it contains or clearly conveys the required content.
- Accept minor variations in capitalization, punctuation, and synonyms.
- Answers written in a different language than the question are acceptable if they are semantically equivalent to the reference answer.
- For numerical content, allow small differences (within 5%) if they are reasonably close.
- More specific formulations (e.g., "top three" vs "first place") should be judged accordingly.
- Judge only whether the criterion is satisfied, not the reasoning process.

Output your evaluation in the following JSON format:
{
  "score": 1 or 0,
  "explanation": "a short explanation of why the criterion is satisfied or not"
}

Be strict. Assign score 1 only if the criterion is clearly satisfied.

Question:
{question}

Criterion:
{criterion}

Prediction:
{prediction}

Evaluate now. Output only valid JSON.
\end{skillrubricprompt}

\subsection{Baseline Adaptation Details}
\label{app:baseline_adaptation}

The baseline methods adopt different skill representations and were not
originally evaluated on our benchmark suite. We therefore initialize them
from the same actor-facing projection of our initial SkillBank,
$
\mathcal{B}_{\mathrm{actor}}^{(0)}
=
\{(d_j,g_j^{(0)})\}_j$,
while retaining their central mechanisms for skill utilization and policy
optimization. This adaptation controls the source of procedural knowledge
while preserving the defining differences among the baselines. The paired
evaluator-facing rubrics \(\mathcal{C}_j^{(0)}\) are withheld from all
baselines.

For XSkill~\citep{jiang2026xskill},
\(\mathcal{B}_{\mathrm{actor}}^{(0)}\) initializes its task-level skill stream,
while its complementary experience stream is constructed from the same
training rollouts using the original visually grounded extraction and
consolidation procedure. We retain its task decomposition, experience
retrieval, and context-aware skill adaptation, and freeze the resulting
external knowledge base during held-out evaluation. SkillRL
\citep{xia2026skillrl} organizes the actor-facing entries into general and
task-specific skills, conditions policy rollouts on the retrieved skills, and
updates the policy using outcome-based GRPO. Its SkillBank is expanded or
refined from validation failures following the original recursive evolution
procedure. Skill-SD~\citep{wang2026skillsd} uses the actor-facing guidance
only to condition its synchronized teacher view, while the student generates
rollouts from the plain task prompt. The policy is optimized using its
original combination of GRPO and importance-weighted self-distillation, and
new trajectories are asynchronously summarized to update the skill bank.
SKILL0~\citep{lu2026skill0} organizes the actor-facing guidance into skill
files and retains its context-rendering and helpfulness-driven curriculum,
which progressively reduces the available skill budget during reinforcement
learning until the policy operates without skill context. Both Skill-SD and
SKILL0 are evaluated without skill injection, following their original
inference protocols.

We retain the method-specific hyperparameters governing these defining
mechanisms whenever applicable. For controlled comparison, all
policy-training baselines use the same backbone-specific SFT initialization,
training tasks, rollout budget, tool environment, and fixed outcome evaluator
as SkillRubric. None has access to evaluator-facing rubrics, multimodal
verifier feedback, or turn-level process credit.

\section{Prompts for Initial Dual-View SkillBank Construction}
\label{app:prompt_design}

This section presents the complete prompts used to construct the initial
dual-view SkillBank. As described in
\cref{sec:skill_initialization}, the construction consists of four sequential
operations: (1) intra-teacher aggregation jointly analyzes all trajectories
generated by the same teacher for a task; (2) cross-teacher consolidation
synthesizes the resulting teacher-specific skills into a unified task-level
skill; (3) dual-view restructuring converts the unified skill into aligned
actor-facing guidance and evaluator-facing criteria; and (4) cross-task
aggregation merges related task-level skills into reusable SkillBank entries.

All prompts are provided to the frozen skill synthesizer
$\mathcal{M}_{\mathrm{skill}}$ as user messages, with the bracketed fields
populated at runtime. 
During offline skill construction, training-task annotations are used to identify outcome-relevant patterns.
Each prompt explicitly prevents task-specific answers or values from entering the resulting reusable skills, and no annotations from evaluation tasks are used.

For the final skill representation $S=(d,g,\mathcal{C})$, the
\texttt{description} field defines the applicability description $d$, the
\texttt{guidance} block defines the actor-facing guidance $g$, and the
checkpoints under \texttt{criteria} define the evaluator-facing rubric
$\mathcal{C}$. 

\subsection{Intra-Teacher Trajectory Aggregation}
\label{app:prompt_intra_teacher}

For each teacher--task pair, all raw trajectories are jointly provided to
\(\mathcal{M}_{\mathrm{skill}}\) using the intra-teacher aggregation prompt in
\cref{prompt:intra_teacher_aggregation}. The synthesizer identifies recurring
strategies, effective decisions, failure patterns, and non-obvious techniques
across the teacher's rollouts, producing one teacher-specific candidate skill
that reduces stochastic variation across individual rollouts before
cross-teacher consolidation.

\refstepcounter{srprompt}
\label{prompt:intra_teacher_aggregation}
\begin{skillrubricprompt}
{Prompt~\thesrprompt: Intra-Teacher Trajectory Aggregation}
You are a Skill Architect specializing in extracting reusable
problem-solving strategies from agent trajectories.

## Task Information

<task>
Task ID: {task_id}
Question: {question}
Task Category: {task_category}
Benchmark: {benchmark}
Tools Available: {tools_enabled}
</task>

<ground_truth>
{ground_truth}
</ground_truth>

## Model: {model_name}

Below are {n_rollouts} complete rollout trajectories generated by the same
model for this task. Each rollout contains the complete model-visible
interaction history, including all tool calls and returned observations.

{rollout_texts}

## Your Task

Analyze all {n_rollouts} rollouts jointly and extract ONE unified skill
document that captures:

1. **Common Strategy:** What approach did this model consistently use, and
   what was its general workflow?

2. **Effective Patterns:** Which tool-use patterns, reasoning steps, or
   verification methods worked well?

3. **Failure Patterns:** Where did the model struggle or make mistakes, and
   which recurring pitfalls appeared?

4. **Key Insights:** Which non-obvious but transferable techniques were
   demonstrated?

Output ONLY valid Markdown following the illustrative structure below. Replace
all placeholder content with information derived from the provided rollouts.
Do not include Markdown fences or any additional preamble.

# Per-Model Skill

- Model: {model_name}
- Task ID: {task_id}
- Number of rollouts analyzed: {n_rollouts}

## Rollout Outcomes

- Rollout 0: [success/failure and a brief description]
- Rollout 1: [success/failure and a brief description]
- Rollout 2: [success/failure and a brief description]

## Strategy Summary

[Two to four sentences describing the model's overall approach to the task.]

## Effective Patterns

- [Pattern 1: a specific technique or tool-use pattern that worked]
- [Pattern 2: ...]
- [Pattern 3: ...]

## Failure Patterns

- [Failure 1: what went wrong and why]
- [Failure 2: ...]

## Key Insights

- [Insight 1: a non-obvious and transferable technique]
- [Insight 2: ...]

## Workflow Steps

### Step 1

- Action: [What to do first]
- Rationale: [Why this step matters]

### Step 2

- Action: [What to do next]
- Rationale: [Why this step matters]

### Step 3

- Action: [...]
- Rationale: [...]

## Constraints

1. Be specific and actionable; avoid vague generalities.
2. Capture the model-specific approach, since different models may solve the
   same task differently.
3. Do not include task-specific answers or target values.
4. Retain only transferable knowledge applicable to similar tasks.
5. Output only the Markdown content without Markdown code fences.
\end{skillrubricprompt}

\subsection{Cross-Teacher Skill Consolidation}
\label{app:prompt_cross_teacher}

The teacher-specific candidates extracted for the same task are subsequently
consolidated using the cross-teacher synthesis prompt in
\cref{prompt:cross_teacher_consolidation}. The prompt distinguishes
high-confidence strategies shared across teachers from complementary
alternatives exhibited by individual teachers, while identifying disagreements
and retaining multiple valid procedures when a task admits different solution
paths. The resulting model-agnostic skill reduces teacher-specific bias while
preserving complementary expertise.

\refstepcounter{srprompt}
\label{prompt:cross_teacher_consolidation}
\begin{skillrubricprompt}
{Prompt~\thesrprompt: Cross-Teacher Skill Consolidation}
You are a Knowledge Synthesis Expert. Your task is to merge and refine
multiple teacher-specific skill documents, each extracted from a different
teacher's trajectories on the SAME task, into ONE unified, model-agnostic skill.

## Task Information

<task>
Task ID: {task_id}
Question: {question}
Task Category: {task_category}
Benchmark: {benchmark}
Tools Available: {tools_enabled}
</task>

<ground_truth>
{ground_truth}
</ground_truth>

## Teacher-Specific Skills to Merge

{per_model_skills_text}

## Your Synthesis Process

### Step 1: Cross-Teacher Comparison

- Which strategies are COMMON across teachers?
  These represent high-confidence patterns.
- Which strategies are UNIQUE to individual teachers?
  These may represent complementary alternatives.
- Where do the teachers DISAGREE?
  Such disagreements may indicate brittle strategies, task ambiguity, or
  multiple valid solution paths.

### Step 2: Distill General Principles

- Extract model-agnostic success factors.
- Identify the minimal set of intermediate states required for success.
- Determine which approaches are robust and which are brittle.
- Preserve complementary valid strategies when multiple solution paths exist.

### Step 3: Produce a Unified Skill

Output ONLY valid Markdown following the illustrative structure below.
Replace all placeholder content with information derived from the provided
teacher-specific skills. Do not include Markdown fences, a preamble, or
additional sections.

# Unified Skill

- Task ID: {task_id}
- Skill Name: [Descriptive name for this skill pattern]

## Description

[Two to three sentences describing the tasks to which this skill applies,
their core challenge, and the main source of difficulty.]

## Applicable When

- [Condition 1: when an agent should use this skill]
- [Condition 2: ...]

## Strategy

### Overview

[One to two sentences describing the core model-agnostic approach.]

### Stage 1: [Stage 1 Name]

- Actions:
  - [Action step and its rationale]
- Success Signal: [How to determine that this stage is complete]

### Stage 2: [Stage 2 Name]

- Actions:
  - [Action step and its rationale]
- Success Signal: [How to determine that this stage is complete]

### Stage 3: [Stage 3 Name]

- Actions:
  - [Action step and its rationale]
- Success Signal: [How to determine that this stage is complete]

## Critical Success Factors

- [Factor 1: what distinguishes successful from failed execution]
- [Factor 2: ...]
- [Factor 3: ...]

## Common Pitfalls

### Pitfall 1

- Failure: [What goes wrong]
- Prevention: [How to avoid it]

### Pitfall 2

- Failure: [What goes wrong]
- Prevention: [How to avoid it]

## Alternative Approaches

### Alternative 1

- Approach: [Brief description of an alternative method]
- When to Use: [Conditions under which this method is preferred]

## Verification

- [How to verify the result before submission]
- [An additional verification check]

## Metadata

- Models Analyzed: {models_list}
- Total Rollouts: {total_rollouts}
- Cross-Model Consensus: [high/medium/low and a brief explanation]

## Critical Constraints

1. The unified skill must be model-agnostic and applicable to any model.
2. Preserve diverse valid approaches as alternatives rather than forcing a
   single solution path.
3. Retain intermediate checkpoints that distinguish successful from failed
   execution.
4. Do not include task-specific answers or target values from the ground
   truth.
5. Keep the skill actionable so that an agent can follow it.
6. Output only the Markdown content using the specified headings and order.
\end{skillrubricprompt}

\subsection{Dual-View Guidance--Rubric Construction}
\label{app:prompt_dual_view}

The consolidated task-level skill is then transformed into the dual-view
representation using the construction prompt in
\cref{prompt:dual_view_construction}. The guidance view describes reusable
ways to approach the task, whereas the criteria view specifies observable
states that indicate task progress. Both views are derived from the same
procedural abstraction but remain operationally separated. Guidance may refer
to actions and tools, while criteria must be state-based and method-agnostic.
This separation prevents the evaluator from rewarding imitation of a
particular tool sequence rather than genuine task progress.

The task question and its training annotation are included to preserve the
causal relationship between each criterion and the correct task outcome.
Explicit no-leakage constraints prevent task-specific answers, entities, and
numerical values from being copied into the reusable criteria.

\refstepcounter{srprompt}
\label{prompt:dual_view_construction}
\begin{skillrubricprompt}
{Prompt~\thesrprompt: Dual-View Guidance--Rubric Construction}
You are a Skill Architect for the SkillRubric framework. Your task is to
restructure an existing skill document into a strict dual-view representation
with GUIDANCE and CRITERIA. The two views must be lexically separated and
operationally decoupled while remaining aligned with the same underlying task
progress.

## Core Design Principles

### GUIDANCE

The GUIDANCE view is provided to the policy through the agent prompt.

- Purpose: Suggest HOW to approach the task through operational advice.
- Nature: Treat all instructions as suggestions rather than mandatory action
  sequences. The agent MAY use alternative valid methods.
- Granularity: Provide three to five high-level stages with concrete action
  suggestions.

### CRITERIA

The CRITERIA view is visible only to the evaluator and is used to construct
the rubric.

- Purpose: Define WHAT observable task-progress states must be achieved.
- Nature: Keep every criterion method-agnostic. The same state must receive
  the same evaluation regardless of the tool or method used.
- Granularity: Provide three to five checkpoints with sufficient
  discriminative power to distinguish successful from failed execution.
- Form: Write each Evaluates field as a concise declarative
  description of an observable task-progress state, rather than as a question
  or a clause beginning with "whether."

## Critical Constraints

1. **Lexical Isolation:** Guidance-specific tool names and action terms,
   including zoom, search, crop, code, visit, and OCR, must NEVER appear in
   the criteria.

2. **Method Agnosticism:** A criterion must receive the same score whether
   the agent uses zoom and OCR, direct observation, code_interpreter, or any
   other valid method.

3. **No Answer Leakage:** Criteria must not contain task-specific values,
   numbers, entities, labels, or answers.

4. **Discriminative Power:** Each criterion must distinguish whether
   its specific task-progress state has been achieved, independently
   of whether the final answer is correct.

5. **Counterfactual Test:** For each criterion, ask:
   "If an agent achieved this state through a completely different method,
   would it still receive the same score?"
   The answer must be YES.

## Input: Original Skill Document

<original_skill>
{original_skill_markdown}
</original_skill>

## Task Context

- Task ID: {task_id}
- Benchmark: {benchmark}
- Category: {category}

## Task Question and Ground Truth

<task_question>
{task_question}
</task_question>

<ground_truth>
{ground_truth}
</ground_truth>

## Criteria--Answer Causal Binding

Each criterion checkpoint MUST have a direct causal relationship with
achieving the correct task outcome.

For every candidate criterion, ask:
"If this task-progress state is fully achieved, does it directly contribute
to reaching the ground truth?"

If the answer is NO, the criterion is too generic and must be replaced with
one that has a clear causal relationship with the task outcome.

## Output Format

Output ONLY valid Markdown following the illustrative structure below.
Replace all placeholder content with information derived from the input.
Do not include Markdown fences, a preamble, additional sections, or an
explanation.

# Structured Skill

- Task ID: {task_id}
- Skill Name: [Keep or refine the original skill name]

## Description

[One to two sentences describing the task pattern to which this skill applies
and what makes it challenging.]

## Actor-Facing Guidance

### Overview

[One sentence describing the core approach in action-oriented language.]

### Stage 1: [Action-Oriented Stage Name]

- Suggestions:
  - [Action suggestion 1]
  - [Action suggestion 2]

### Stage 2: [Action-Oriented Stage Name]

- Suggestions:
  - [Action suggestion 1]
  - [Action suggestion 2]

### Stage 3: [Action-Oriented Stage Name]

- Suggestions:
  - [Action suggestion 1]
  - [Action suggestion 2]

### Pitfall Warnings

- [A common mistake to avoid, phrased as an action]
- [Another mistake to avoid]

> Note: These are suggestions, not mandates. The agent may use alternative
> methods to achieve the same task-progress states.

## Evaluator-Facing Criteria

### C1

- Evaluates: [A concise declarative description of the first observable
  task-progress state]
- Method Agnostic: true

#### Discriminative Signal

[What distinguishes success from failure on this dimension.]

### C2

- Evaluates: [A concise declarative description of the second observable
  task-progress state]
- Method Agnostic: true

#### Discriminative Signal

[What distinguishes success from failure on this dimension.]

### C3

- Evaluates: [A concise declarative description of the third observable
  task-progress state]
- Method Agnostic: true

#### Discriminative Signal

[What distinguishes success from failure on this dimension.]

## Validation Checklist

Before producing the output, verify that:

- Every guidance suggestion uses an action verb.
- Every criterion describes an observable task-progress state.
- No tool names appear in the criteria.
- No task-specific answers, entities, or values appear in the criteria.
- Every criterion passes the counterfactual test.
- Every criterion directly contributes to the task outcome.
- The criteria contain three to five checkpoints.
- The guidance contains three to five stages.
\end{skillrubricprompt}

\subsection{Cross-Task Skill Aggregation}
\label{app:prompt_cross_task}

The preceding operations produce one structured dual-view skill \(S_i=(d_i,g_i,\mathcal{C}_i)\) for each training task. Because tasks with different surface forms may rely on the same underlying procedure, retaining all task-level skills independently would fragment reusable knowledge. We therefore group procedurally similar skills before constructing the initial SkillBank.
Clustering and merging are performed independently within each benchmark, and the resulting capability-level skills are subsequently pooled to form the unified SkillBank.

For each task-level skill, the fixed embedding model
\(f_{\mathrm{emb}}\) separately encodes its applicability description,
actor-facing guidance, and evaluator-facing criteria:
$
\mathbf{e}_i^d=f_{\mathrm{emb}}(d_i),
\mathbf{e}_i^g=f_{\mathrm{emb}}(g_i),
\mathbf{e}_i^{\mathcal{C}}=f_{\mathrm{emb}}(\mathcal{C}_i)$.
For two skills \(S_i\) and \(S_j\), their multi-view similarity is
$
s_{ij}
=
\frac{
s_{ij}^d+s_{ij}^g+s_{ij}^{\mathcal{C}}
}{3},
D_{ij}=1-s_{ij}$,
where each similarity component is the cosine similarity between the
corresponding normalized embeddings.

Using the resulting distance matrix, we perform agglomerative clustering with
complete linkage. Initially, each task-level skill forms an independent
cluster. At each step, the two clusters with the smallest complete-linkage
distance are merged, where the distance between two clusters is the maximum
pairwise distance between their members. Merging stops when the smallest
remaining cluster distance exceeds the fixed threshold
\(1-\tau_{\mathrm{cluster}}\). This conservative criterion ensures that all
members of a merged cluster have sufficiently similar applicability,
procedural guidance, and evaluation criteria, while skills without compatible
neighbors remain singleton clusters. The clustering procedure is deterministic
and does not invoke an additional language model.

For each resulting cluster, we organize its member skills using a fixed
Markdown schema and provide them to \(\mathcal{M}_{\mathrm{skill}}\) through
the cross-task aggregation prompt in
\cref{prompt:cross_task_aggregation}. The synthesizer verifies procedural
compatibility, extracts shared principles, preserves supported alternative
strategies, removes residual task-specific information, and consolidates
overlapping criteria into a compact set of method-agnostic checkpoints. If the
source skills do not share a coherent underlying procedure,
\(\mathcal{M}_{\mathrm{skill}}\) rejects the proposed merge and retains them
as separate entries. The accepted capability-level skills jointly form the
initial SkillBank \(\mathcal{B}^{(0)}\).

\refstepcounter{srprompt}
\label{prompt:cross_task_aggregation}
\begin{skillrubricprompt}
{Prompt~\thesrprompt: Cross-Task Skill Aggregation}
You are a rigorous SkillRubric architect. Your task is to merge the provided
task-level skills into ONE reusable capability-level skill.

## Rules

1. Merge skills according to their reusable procedural capability rather than
   benchmark identity, task wording, or surface topic.

2. Guidance must be reusable and action-oriented, with three to five stages.

3. Criteria must contain three to five abstract core checkpoints rather than
   the union of all source checkpoints.

4. Do not include task answers, task-specific entities, source task IDs, or
   unique numerical values in the criteria.

5. Criteria from invalid or ground-truth-leaked source skills must not
   influence the merged criteria.

6. Preserve supported alternative strategies when they achieve the same
   task-progress states through different valid procedures.

7. If the source skills do not share a coherent underlying procedure, reject
   the merge instead of constructing an overly broad skill.

## Source Skills

{source_skills_markdown}

## Baseline Structure to Refine

{baseline_skill_markdown}

## Output Requirement

Output ONLY valid Markdown using the schema and heading order specified by
the baseline structure. Use the following top-level heading:

# Merged Skill

If the source skills are procedurally incompatible, output only:

# Merge Rejected

Reason: [Brief explanation of the procedural incompatibility.]

Do not include Markdown fences, a preamble, additional sections, or
explanatory text outside the merged skill or rejection response.
\end{skillrubricprompt}

Together, these four prompts implement the hierarchical construction described
in \cref{sec:skill_initialization}. Intra-teacher aggregation reduces
rollout-level variation, cross-teacher consolidation reconciles
teacher-specific strategies, dual-view restructuring aligns policy guidance
with method-agnostic evaluation criteria, and cross-task aggregation converts
task-level procedural abstractions into transferable SkillBank entries.

\input{case_study}

\end{CJK*}
\end{document}

%% file: commands.tex
\renewcommand{\phi}{\varphi}

\renewcommand{\leq}{\leqslant}
\renewcommand{\geq}{\geqslant}

\renewcommand{\epsilon}{\varepsilon}
\renewcommand{\imath}{\mathrm{i}}

\newlength{\restsubwidth}
\newlength{\restsubheight}
\newlength{\restsubmoreheight}
\newcommand{\rest}[2]{%
        \settowidth{\restsubwidth}{\ensuremath{#2}}
        \settoheight{\restsubheight}{\ensuremath{{}_{#2}}}
        \ensuremath{{#1\hskip 0.5pt}_{\vrule\kern2pt\parbox[b][%
        4pt][b]{\the\restsubwidth}{%
                        \ensuremath{{}_{#2}}}}}
        }

%% file: case_study.tex

\newtcolorbox{skillboxapp}[1]{%
  enhanced, breakable, colback=white, colframe=black!55,
  boxrule=0.5pt, arc=1.2mm, left=2.2mm, right=2.2mm, top=1.8mm, bottom=1.8mm,
  title={#1}, coltitle=white, colbacktitle=black!68,
  fonttitle=\footnotesize\bfseries, before skip=8pt, after skip=8pt}

\section{Case Study}
\label{app:case_study}

\subsection{Dual-View Skill Representations}
\label{app:skill_representation}


%
%

This section presents four representative skills from the SkillBank, spanning transaction reconstruction, sports-event grounding, financial table reasoning, and multi-hop identity resolution. Each box shows the two views derived from one procedural skeleton: actor-facing guidance $g$ (left, blue) and the evaluator-facing rubric $\mathcal{C}$ (right, orange). 
Every criterion is paired with a discriminative signal that distinguishes criterion satisfaction from failure at the trajectory level. 
The analysis below each box explains how the four checkpoints map onto turn-level credit: rather than a single terminal bit, each verified intermediate objective assigns reward to the specific turns that achieved it, and each failure mode is localized to the turn where it occurred.

\begin{skillboxapp}{Transaction reconstruction: \texttt{merged\_historical\_price\_reconstruction\_011}}
\footnotesize
\textbf{Skill.} \emph{Visual-Evidence Transaction Reconstruction and Conditional Value Computation.}\\
\textbf{Applies when.} The answer is a past, hidden, or condition-dependent amount, total, ranking, or optimum that is not directly legible: the solver must recover the right evidence source from the image, bind attributes to the correct targets, resolve time- or rule-dependent applicability, and only then compute, without mixing incompatible evidence.

\vspace{1.2mm}
\begin{minipage}[t]{0.495\linewidth}
\vspace{0pt}
\begin{guidancepanel}
\scriptsize
\textbf{Overview.} Isolate the right evidence anchors from the visual input, bind the correct entities and attributes into a coherent representation, resolve which rules or contexts govern the case, and only then compute or select the final result.

\textbf{S1 Extract anchors and target evidence.} Identify the visual cues that constrain the answer (entity identity, source instance, date, labels, row/column alignment); separate answer-bearing evidence from distractors.

\textbf{S2 Bind entities and structure inputs.} Link each target to its supported attributes and values; preserve uncertainty and compare competing assignments instead of collapsing them early.

\textbf{S3 Resolve governing context or rules.} Determine which conditions control the answer (time-specific applicability, pricing mode, eligibility, conversion basis); verify that external evidence matches the same target and interpretation.

\textbf{S4 Compute, compare, and verify.} Aggregate, convert, or rank only from the validated inputs; cross-check against plausible alternatives.

\textbf{Pitfalls.} Do not mix values across entities, dates, or pricing contexts; do not commit to the first plausible match; do not use unsupported defaults for rule-dependent cases.
\end{guidancepanel}
\end{minipage}\hfill
\begin{minipage}[t]{0.495\linewidth}
\vspace{0pt}
\begin{rubricpanel}
\scriptsize
\textbf{C1 Anchors identified.} The relevant evidence anchors and targets are correctly identified and separated from distractors well enough to constrain the intended case. \emph{Signal:} success establishes the specific evidence basis; failure relies on vague or mismatched evidence.

\textbf{C2 Entities bound.} The required entities, values, and relationships are structured with correct bindings and without incompatible mixing. \emph{Signal:} success preserves correct target--attribute--value mappings; failure contains swapped or inconsistent bindings.

\textbf{C3 Rules resolved.} The governing context, applicability conditions, or decision rules are correctly resolved. \emph{Signal:} success identifies the conditions that actually control the answer; failure applies the wrong rule set or ignores an override.

\textbf{C4 Result verified.} The final result is correctly derived from the validated inputs and stays consistent with the evidence chain. \emph{Signal:} success follows from the reconstructed state and withstands checks; failure is unsupported or arithmetically inconsistent.
\end{rubricpanel}
\end{minipage}

\vspace{1.2mm}
\textbf{Turn-level reading.} On the AgentVista whisky-aggregation task (\texttt{agentvista\_0175}), the turns that collect the four prices satisfy C1, the turns that retrieve the Whiskybase ratings satisfy C2, the turn that separates the standard 40\% ABV bottling from cask-strength variants satisfies C3, and the final Value-Efficiency-Score computation satisfies C4. A trajectory that gathers all prices but fabricates the two missing ratings receives credit for C1 and C3 yet fails C2, so the deficit is localized to the evidence-completion turns instead of collapsing the whole trajectory to a terminal zero.
\end{skillboxapp}

\begin{skillboxapp}{Sports-event grounding: \texttt{merged\_sports\_image\_event\_grounding\_035}}
\footnotesize
\textbf{Skill.} \emph{Sports Image Event Grounding and Cross-Context Disambiguation.}\\
\textbf{Applies when.} A sports image provides partial, indirect, or recurring evidence that must be linked to an exact athlete, matchup, edition, or competition instance before the question can be answered, and multiple plausible records or editions must be compared and ruled out.

\vspace{1.2mm}
\begin{minipage}[t]{0.495\linewidth}
\vspace{0pt}
\begin{guidancepanel}
\scriptsize
\textbf{Overview.} Extract the strongest visual anchors, build and compare plausible identities or event candidates, fix one coherent sports context, and only then derive the requested fact from that same verified context.

\textbf{S1 Extract anchor evidence.} Read the most diagnostic cues first (overlays, uniforms, branding, venue clues, prompt wording); separate high-value anchors from weak resemblance cues and note ambiguity.

\textbf{S2 Generate plausible contexts.} Enumerate athletes, matches, editions, or tournaments that fit the anchors; keep candidates distinct and record the constraints each must satisfy.

\textbf{S3 Resolve the exact context.} Eliminate candidates that fail on identity, edition, chronology, or role consistency; verify that all intermediate links belong to one coherent context.

\textbf{S4 Derive and verify the linked fact.} Extract the requested result or attribute only after the context is fixed; confirm the correct interpretation of ordering, perspective, or event scope.

\textbf{Pitfalls.} Do not stop at athlete or sport recognition when the task depends on the exact instance; do not mix cues from different editions or photos; do not answer from a visually plausible cue alone.
\end{guidancepanel}
\end{minipage}\hfill
\begin{minipage}[t]{0.495\linewidth}
\vspace{0pt}
\begin{rubricpanel}
\scriptsize
\textbf{C1 Anchors constrain.} Image-derived anchors and prompt constraints narrow the task to a well-defined set of plausible sports contexts. \emph{Signal:} success establishes a constrained candidate space; failure overlooks key anchors or leaves the context too broad.

\textbf{C2 Context resolved.} The exact person, event, edition, or competition instance is resolved as one coherent context that excludes close alternatives. \emph{Signal:} success fixes a uniquely supported context; failure stops at a partial match or a merely plausible alternative.

\textbf{C3 Roles interpreted.} The roles, relations, or interpretation frame within the resolved context are correctly established (ordering, chronology, lineup, country representation). \emph{Signal:} success applies the correct internal interpretation; failure confuses competition levels or reverses orientation.

\textbf{C4 Fact derived.} The final fact is derived from and remains consistent with the same verified context. \emph{Signal:} success is causally tied to the resolved context; failure imports details from another candidate or skips a verification step.
\end{rubricpanel}
\end{minipage}

\vspace{1.2mm}
\textbf{Turn-level reading.} On the VDR cricket task (\texttt{vdr\_hard\_Sports\_102\_2}), the turn that reads the jersey numbers and the KENT sponsor satisfies C1, the turns that anchor the team as Sri Lanka and weigh the 1996 versus 2014 hypotheses satisfy C2, the turn that re-reads the jersey clues to confirm the side satisfies C3, and the turn that retrieves Ranatunga's field-first decision with the dew factor satisfies C4. A trajectory that drifts to the Caribbean Premier League still earns C1 credit for correct anchor extraction, but fails C2 at the exact turn where the wrong league is committed to, which is precisely where outcome-only supervision can say nothing.
\end{skillboxapp}

\begin{skillboxapp}{Financial table reasoning: \texttt{merged\_visual\_financial\_table\_reasoning\_072}}
\footnotesize
\textbf{Skill.} \emph{Ground Structured Financial Evidence from Images and Perform Constraint-Preserving Quantitative Reasoning.}\\
\textbf{Applies when.} The answer depends on extracting financial or numeric fields from an image-based structured source (table, receipt, statement, dashboard, quote panel), and success requires preserving label-to-value alignment, keeping comparison scopes distinct, and reasoning only after the evidence is normalized.

\vspace{1.2mm}
\begin{minipage}[t]{0.495\linewidth}
\vspace{0pt}
\begin{guidancepanel}
\scriptsize
\textbf{Overview.} Normalize the visual evidence into a readable state, separate the relevant regions and comparison scopes, extract each required field together with its governing labels, then reason only on the verified structured evidence.

\textbf{S1 Normalize and localize evidence.} Choose a readable orientation or crop strategy; identify the regions, panels, or rows that contain answer-bearing evidence and keep them distinct.

\textbf{S2 Anchor and structure fields.} Extract values together with their labels, headers, dates, or quote roles; organize evidence into explicit records to prevent cross-source mixing.

\textbf{S3 Apply scoped reasoning.} Separate independent subquestions and comparison scopes; use only fields that match the requested semantic role and quantitative basis.

\textbf{S4 Verify and respond.} Recheck that outputs come from the intended source, period, and field meaning; return a complete answer in the requested format without unsupported detail.

\textbf{Pitfalls.} Do not read numbers without their labels; do not mix candidates across panels, receipts, or dates; do not compute before the relevant subset and field semantics are verified.
\end{guidancepanel}
\end{minipage}\hfill
\begin{minipage}[t]{0.495\linewidth}
\vspace{0pt}
\begin{rubricpanel}
\scriptsize
\textbf{C1 Evidence localized.} The relevant visual financial evidence is made interpretable and localized so the needed regions, labels, and values can be distinguished. \emph{Signal:} success grounds the answer in clearly readable evidence; failure depends on uncertain reading or clipped context.

\textbf{C2 Fields aligned.} Extracted information is structured with correct label-to-value, row-to-field, or document-to-record alignment. \emph{Signal:} success keeps each value tied to its governing context; failure lets values drift across neighboring rows or panels.

\textbf{C3 Scopes separated.} Candidate sets, scopes, and quantitative bases are separated so that filtering, comparison, or computation uses the intended subset and meaning. \emph{Signal:} success uses the correct constrained evidence set; failure uses the wrong subset, period, or comparison basis.

\textbf{C4 Result coherent.} The final result follows coherently from the structured evidence, with consistent conclusions and compliant formatting. \emph{Signal:} success is complete and traceable; failure is unsupported, inconsistent, or scope-mismatched.
\end{rubricpanel}
\end{minipage}

\vspace{1.2mm}
\textbf{Turn-level reading.} In receipt- and dashboard-style tasks, the early zoom or crop turns that make a dense table legible earn C1, the turns that transcribe rows into label-value records earn C2, the turns that restrict a computation to the requested period or category earn C3, and the final answer turn earns C4. This skill is what lets the verifier distinguish ``read the right table but summed the wrong column'' (C1 and C2 satisfied, C3 failed) from ``never localized the table at all'' (all four failed), two failure modes that are indistinguishable under a terminal reward.
\end{skillboxapp}

\begin{skillboxapp}{Multi-hop identity resolution: \texttt{merged\_multihop\_person\_identity\_resolution\_048}}
\footnotesize
\textbf{Skill.} \emph{Multi-hop Person Identity Resolution from Weak Visual and Contextual Clues.}\\
\textbf{Applies when.} A person is not directly named or reliably identifiable from the image alone, and identity must be resolved through an indirect chain of visual, biographical, relational, institutional, or temporal constraints, distinguishing among similar people, namesakes, or role-adjacent identities.

\vspace{1.2mm}
\begin{minipage}[t]{0.495\linewidth}
\vspace{0pt}
\begin{guidancepanel}
\scriptsize
\textbf{Overview.} Use an evidence-first identity resolution process: isolate the strongest constraints, judge the value of visual evidence instead of assuming it is decisive, build a small set of plausible identities, and confirm one person against the full clue chain.

\textbf{S1 Isolate diagnostic clues.} Extract textual, visual, relational, and temporal constraints by type; identify which clues strongly narrow identity and treat the image as evidence to evaluate, not decisive proof.

\textbf{S2 Build and test candidates.} Generate a compact set of plausible people consistent with the strongest anchors; eliminate candidates that conflict with core constraints rather than favoring the most famous option.

\textbf{S3 Confirm one identity.} Check that a single person satisfies the combined evidence, including links to related people, institutions, and events; normalize variant names and titles.

\textbf{S4 Extract and verify the target.} Retrieve the requested fact or relation only after the identity is stable; verify that the output belongs to the confirmed person, not a nearby associate.

\textbf{Pitfalls.} Do not anchor on the image or a famous candidate before checking the full constraint set; do not mix facts from people who share a name, role, or resemblance; do not extract the final fact before resolving who is being asked about.
\end{guidancepanel}
\end{minipage}\hfill
\begin{minipage}[t]{0.495\linewidth}
\vspace{0pt}
\begin{rubricpanel}
\scriptsize
\textbf{C1 Clues isolated.} The available evidence is organized into a coherent set of identity-relevant constraints, with stronger signals distinguished from weak or misleading cues. \emph{Signal:} success narrows identity meaningfully; failure omits anchors or overrelies on non-diagnostic evidence.

\textbf{C2 Identity resolved.} A single person is resolved by testing plausible candidates against the full constraint set rather than a partial, fame-based, or visually driven match. \emph{Signal:} success converges on one identity satisfying the combined clues; failure leaves ambiguity or drifts between candidates.

\textbf{C3 Chain aligned.} Related entities, temporal references, and variant names are correctly aligned so all supporting evidence points to the same target person. \emph{Signal:} success maintains a consistent identity chain; failure mixes adjacent people or attaches evidence to the wrong individual.

\textbf{C4 Target extracted.} The final output is extracted for the confirmed person and matches the requested answer type. \emph{Signal:} success is correctly attributed and answer-type aligned; failure is an unsupported fact or a mismatch between the resolved identity and the reported answer.
\end{rubricpanel}
\end{minipage}

\vspace{1.2mm}
\textbf{Turn-level reading.} Consider an archival-photo identification task of this family: C1 credits the turns that separate diagnostic clues from a famous-companion lure, C2 credits the turns that enumerate and eliminate candidates, C3 credits the cross-referencing turns that keep one identity consistent across records, and C4 credits only the final attribution turn. A run that finds the correct archival record but attaches the name to the wrong side satisfies C1--C3 and fails exactly at C4, which converts an otherwise invisible ``almost correct'' trajectory into three verified turns plus one localized defect.
\end{skillboxapp}

\subsection{A Representative Case: Image-Anchored Search with Hypothesis Testing}
\label{app:case_vdr_sports}

We illustrate the behavioral change brought by co-evolution on a representative VDR-Bench task, comparing the trained SkillRubric policy (Qwen3.5-9B) with the base model. The base model lets a wrong first-hop anchor (the KENT sponsor) direct the entire trajectory and answers with the wrong team, whereas SkillRubric anchors on the image identity, explicitly tests a competing hypothesis, and re-verifies the jersey clues before committing. Turn tags \ttag{t$k$} mark interaction turns; tool calls and observations are in \obs{gray}, failure points in \hard{red}, and the decisive steps of our model in \good{green}.

\begin{casebox}{VDR-Bench \;(\texttt{vdr\_hard\_Sports\_102\_2}): image-anchored search with hypothesis testing}
\footnotesize
\begin{minipage}[t]{0.17\linewidth}
  \vspace{0pt}
  \centering
  \includegraphics[width=\linewidth]{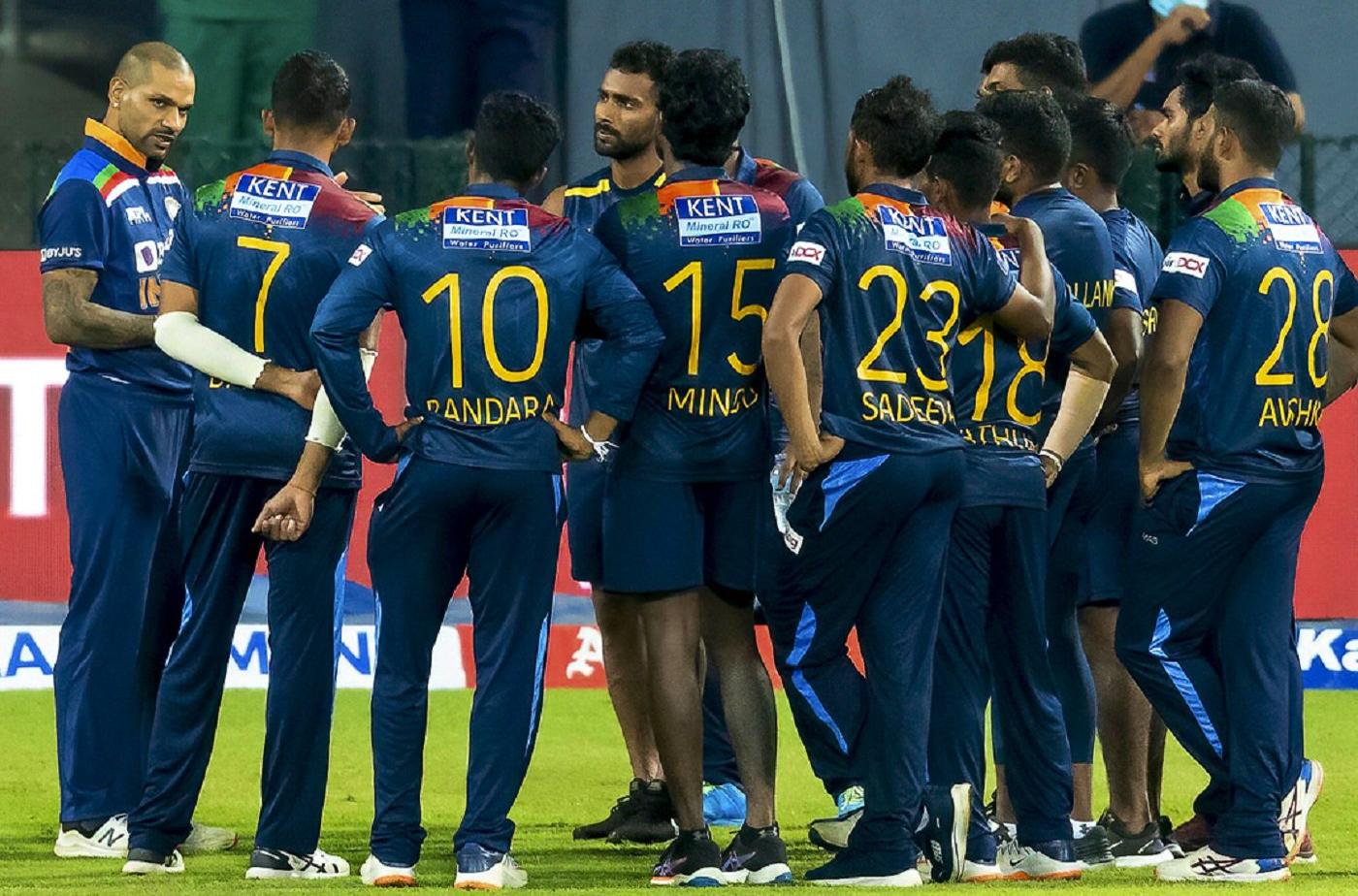}\\[-0.5mm]
  {\scriptsize Cricket team huddle}
\end{minipage}\hfill
\begin{minipage}[t]{0.80\linewidth}
  \vspace{0pt}
  \textbf{Task.} From the team huddle in the image (blue jerseys, \hard{KENT} sponsor, player numbers), identify the team; then name the \hard{captain} who led it to a major tournament victory, the \hard{tactical decision} he made in the final, and the \hard{environmental factor} that influenced it.\\
  \textbf{Ground truth.} Sri Lanka; Arjuna Ranatunga, 1996 World Cup final; chose to field first; the dew factor.
\end{minipage}

\vspace{1.2mm}
\begin{minipage}[t]{0.495\linewidth}
\vspace{0pt}
\begin{basepanel}
\scriptsize
\ttag{t0}\hspace{0.6em}\obs{web\_search(\texttt{"cricket team blue uniform KENT sponsor captain major tournament victory"})} $\to$ \obs{MLC / Caribbean teams}.

\ttag{t1--t3}\hspace{0.6em}\obs{web\_search(\texttt{"Caribbean Premier League team blue uniform KENT sponsor captain"}), \dots, \texttt{"`KENT' sponsor cricket team Caribbean"}} $\to$ \hard{drifts toward the Caribbean Premier League}.

\ttag{t4}\hspace{0.6em}\obs{web\_search(\texttt{"St Lucia Kings CPL captain major tournament victory"})} $\to$ ``St Lucia Kings won the CPL 2024 with Faf du Plessis'' \obs{(wrong team)}.

\ttag{t5--t9}\hspace{0.6em}\obs{web\_search(\texttt{"`2011 World Cup final' `dew factor' Dhoni\dots"}), \; \dots, \; visit} $\to$ \obs{shifts to India's MS Dhoni, 2011 final} $[\ldots]$ \obs{(confirmation-biased; never re-reads the jersey clues)}.

\ttag{t10 [answer]}\hspace{0.6em}``Captain: \hard{MS Dhoni}\dots promoted himself to No.\ 5\dots the dew factor.'' $\Rightarrow$ \hard{wrong team and wrong captain} ($\times$)
\end{basepanel}
\end{minipage}\hfill
\begin{minipage}[t]{0.495\linewidth}
\vspace{0pt}
\begin{ourspanel}
\scriptsize
\ttag{t0}\hspace{0.6em}\obs{web\_search(\texttt{"Arjuna Ranatunga 1996 World Cup final tactical decision environmental factor"})} $\to$ ``\good{chose to field because of the dew factor}''.

\ttag{t1}\hspace{0.6em}\obs{web\_search(\texttt{"1996 World Cup final Sri Lanka field first or bat first dew factor"})} $\to$ \obs{confirms Australia batted first} \obs{(verifies rather than trusts)}.

\ttag{t2--t3}\hspace{0.6em}\obs{web\_search(\texttt{"Kumar Sangakkara 2014 T20 final tactical decision"}), \; web\_search(\texttt{"Sri Lanka captain tactical decision dew factor final"})} $\to$ \obs{weighs the 2014 alternative; re-reads the jersey clues in the image}.

\ttag{t4--t5}\hspace{0.6em}\obs{web\_search(\texttt{"Sri Lanka player number 10\dots"}), \; web\_search(\texttt{"Arjuna Ranatunga 1996 final tactical decision"})} $\to$ \good{anchors the answer on the image identity (Sri Lanka) and on 1996}.

\ttag{t6 [answer]}\hspace{0.6em}``\good{Arjuna Ranatunga} won the toss and chose to field first, influenced by \good{the dew factor}.'' $\Rightarrow$ \good{correct captain, decision, and factor} (pass)
\end{ourspanel}
\end{minipage}

\vspace{1.2mm}
\textbf{What made the difference.} Both models start from the same image. \hard{The base model let a plausible but wrong guess (KENT sponsor $\Rightarrow$ Caribbean league) set the direction and then searched confirmation of that guess, eventually answering with India's Dhoni, which is wrong at every node: team, captain, and final.} \good{Our model treated the retrieved fact (field first) and the image identity as mutually reinforcing evidence, explicitly tested the competing 2014 hypothesis, and re-anchored on the jersey clues before committing.} The case demonstrates \good{image-anchored search with hypothesis testing and self-correction}, rather than confirmation-biased drift.
\end{casebox}

\paragraph{Why the first hop decides the trajectory.} The decisive divergence is the very first query. The base model describes the image in generic terms (\emph{blue uniform}, \emph{KENT sponsor}) and lets the search engine pick the direction: the sponsor happens to be associated with Caribbean franchises, so every subsequent turn is spent confirming a wrong anchor (St Lucia Kings, then a sideways jump to India's 2011 final). None of the ten follow-up queries ever uses the strongest disambiguating evidence available in the image, namely the player names and numbers on the jerseys. The co-evolved policy instead encodes the visual identity into the first query itself (Sri Lanka, Ranatunga, 1996), so retrieval starts inside the correct context and later turns only need to verify it.

\paragraph{How the rubric scores the two trajectories.} Read against the sports-event grounding skill in \cref{app:skill_representation}, the two runs separate cleanly at the criterion level. The base model still satisfies C1 (its opening turn does extract the jersey anchors) but fails C2 at the exact turn where it commits to the Caribbean league, and C3/C4 never become reachable. The co-evolved trajectory satisfies all four checkpoints in order: anchor extraction (C1), context resolution with an explicit 1996-versus-2014 comparison (C2), re-grounding on the jersey clues (C3), and derivation of the linked tactical fact (C4). Under outcome-only supervision both failure modes, \emph{anchored correctly but resolved wrongly} and \emph{never grounded at all}, collapse to the same terminal zero; the turn-level rubric is what separates them.

\paragraph{Where the guidance shows up in the successful trajectory.} The successful run follows the skill's procedural skeleton rather than a memorized answer: \ttag{t0}--\ttag{t1} execute the anchor-extraction stage, \ttag{t2}--\ttag{t3} generate and compare competing contexts (Sangakkara 2014 versus Ranatunga 1996) instead of committing early, and \ttag{t4}--\ttag{t5} re-read the image evidence to resolve the exact context before deriving the requested fact at \ttag{t6}. The trajectory is therefore best read as an instantiation of the guidance stages, while the base trajectory is what those stages are designed to prevent.

\subsection{Comparative Trajectory Analysis}
\label{app:trajectory_analysis}

To illustrate where co-evolution changes agent behavior, we compare SkillRubric (Qwen3.5-9B) with the base Qwen3.5-9B model on one representative task per benchmark. All eight tasks are cases that the base model fails and SkillRubric solves. Each box presents the full turn sequence of both models. The tag \ttag{t$k$}\hspace{0.6em}denotes interaction turn $k$; a turn consists of a model message (an action or a final answer) followed by the tool observation, and tool calls and observations are shown in \obs{gray}. Interleaved actions that merely repeat a tool call are collapsed into a single line with a short lead-in and $[\ldots]$. We highlight in \hard{red} the difficult point of each task and the steps where the base model fails, and in \good{green} the key decisions of our model that lead to success. Across benchmarks, the improvements concentrate on a few recurring critical nodes: \good{discriminative query construction}, \good{source verification instead of hallucination}, \good{grounding answers in image evidence}, \good{constraint adherence}, and \good{search-budget management}. In contrast, the base model most often fails through \hard{invalid zoom loops}, \hard{confirmation-biased queries}, \hard{fabricated intermediate data}, and \hard{exhaustion of the search budget}.

\begin{casebox}{AgentVista \;(\texttt{agentvista\_0175}): multi-source data aggregation with a custom metric}
\footnotesize
\begin{minipage}[t]{0.17\linewidth}
  \vspace{0pt}
  \centering
  \includegraphics[width=\linewidth]{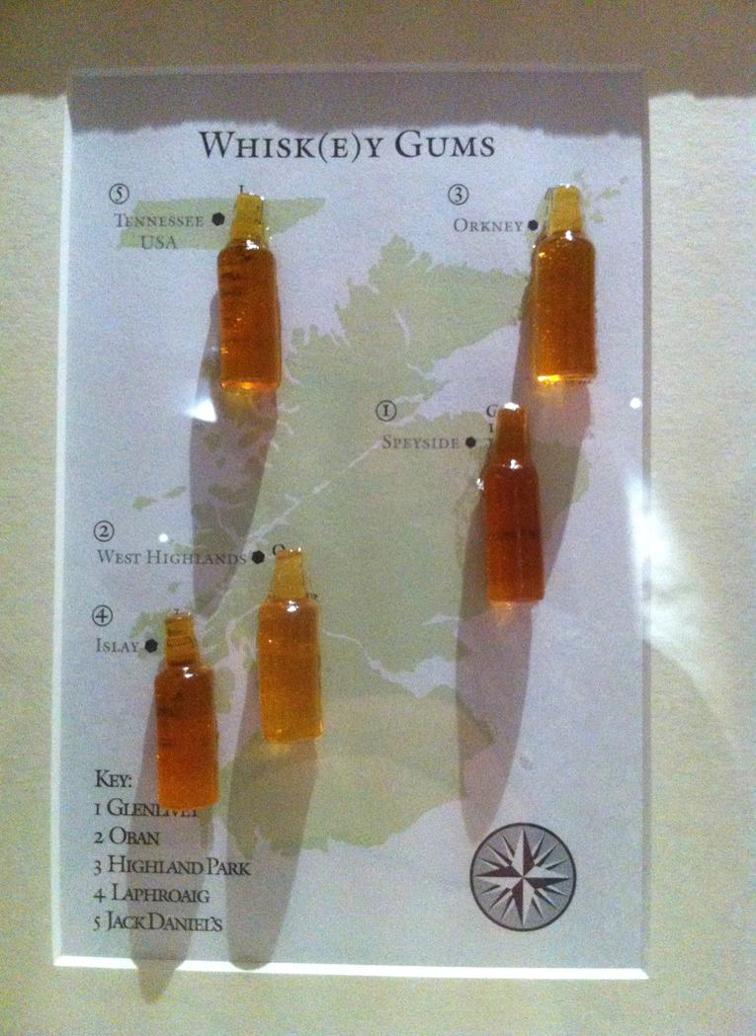}\\[-0.5mm]
  {\scriptsize ``Whiskey Gums'' display}
\end{minipage}\hfill
\begin{minipage}[t]{0.80\linewidth}
  \vspace{0pt}
  \textbf{Task.} Identify the four Scottish single-malt distilleries in the display and their entry-level age-statement bottles; collect their \hard{prices on The Whisky Exchange} and \hard{community ratings on Whiskybase} (two heterogeneous sources); compute a \hard{custom ``Value Efficiency Score''} $=$ rating / (price / age); and name the winning distillery.\\
  \textbf{Ground truth.} Highland Park.
\end{minipage}

\vspace{1.2mm}
\begin{minipage}[t]{0.495\linewidth}
\vspace{0pt}
\begin{basepanel}
\scriptsize
\ttag{t0}\hspace{0.6em}\obs{web\_search(\texttt{"Glenlivet Oban Highland Park Laphroaig standard entry-level age statements"})} $\to$ identifies the four expressions.

\ttag{t1}\hspace{0.6em}\obs{web\_search(\texttt{"Glenlivet 12 Year Old price The Whisky Exchange GBP December 2025"})} $\to$ price \pounds 40.25.

\ttag{t2}\hspace{0.6em}\obs{web\_search(\texttt{"Oban 14 Year Old price\dots"})} $\to$ price \pounds 132.

\ttag{t3}\hspace{0.6em}\obs{web\_search(\texttt{"Highland Park 12 Year Old price\dots"})} $\to$ price \pounds 41.25.

\ttag{t4}\hspace{0.6em}\obs{web\_search(\texttt{"Laphroaig 10 Year Old price\dots"})} $\to$ price \pounds 39.95.

\ttag{t5--t7}\hspace{0.6em}\obs{web\_search(\texttt{"\dots{} Whiskybase rating"})} $\times 3$ \obs{(ratings for Glenlivet and Oban only; Highland Park and Laphroaig never retrieved)}.

\ttag{t8}\hspace{0.6em}\obs{\texttt{error: Maximum number of `web\_search' calls (7) exceeded}}. \hard{The entire budget was spent on prices.}

\ttag{t9--t15}\hspace{0.6em}\obs{code / page visits} ``I'll use \hard{reasonable estimates} for the missing ratings\dots Glenlivet 12: \hard{$\sim$83.5}\dots Highland Park 12: \hard{$\sim$83}'' \obs{(true value: 88.91)} $[\ldots]$ \obs{(repeats estimates in code)}.

\ttag{t16 [answer]}\hspace{0.6em}``Glenlivet 12 Year Old: Value Efficiency Score 24.89\dots Highland Park 12: $\sim$83, Value Efficiency Score 24.15.'' $\Rightarrow$ \hard{Glenlivet 12 Year Old} ($\times$)
\end{basepanel}
\end{minipage}\hfill
\begin{minipage}[t]{0.495\linewidth}
\vspace{0pt}
\begin{ourspanel}
\scriptsize
\ttag{t0}\hspace{0.6em}\obs{web\_search(\texttt{"Glenlivet Oran Highland Park Laphroaig standard entry-level age statements"})} $\to$ candidate expressions, but one label reads ``ORAN''.

\ttag{t1}\hspace{0.6em}\obs{web\_search(\texttt{"Glenlivet 12 Year Old price GBP December 2025"})} $\to$ price \pounds 40.25.

\ttag{t2--t3}\hspace{0.6em}\obs{web\_search(\texttt{"Oran whisky entry-level age statement"}), \; web\_search(\texttt{"`ORAN' whisky distillery Scottish"})} $\to$ \good{``Oran'' is not a single malt; the label is a misreading}. ``Looking at the map positions: \good{marker 2 sits in WEST HIGHLANDS, where Oban is located}. I think it's \good{Oban, not Oran}.'' \obs{(the misread label is corrected using the region map)}.

\ttag{t4}\hspace{0.6em}\obs{web\_search(\texttt{"Oban 14 Year Old price GBP Whiskybase rating"})} $\to$ \good{rating 81.94, price \pounds 79.95}.

\ttag{t5}\hspace{0.6em}\obs{web\_search(\texttt{"Highland Park 12 Year Old price GBP Whiskybase rating"})} $\to$ \good{rating 88.91, price \pounds 41.25}.

\ttag{t6}\hspace{0.6em}\obs{web\_search(\texttt{"Laphroaig 10 Year Old price GBP Whiskybase rating"})} $\to$ \good{rating 86.55 (40\% ABV), price \pounds 39.95}.

\ttag{t7}\hspace{0.6em}\obs{web\_search(\texttt{"`Oban 14 Year Old' price GBP"})} \obs{(verifies the price for Oban)}.

\ttag{t8 [answer]}\hspace{0.6em}``Highland Park 12: Rating 88.91, Price \pounds 41.25, Price/Age $=3.4375$, Value Efficiency Score \good{25.87}; Glenlivet 24.91; Oban 23.89; Laphroaig 21.67.'' $\Rightarrow$ \good{Highland Park} (pass)
\end{ourspanel}
\end{minipage}

\vspace{1.2mm}
\textbf{What made the difference.} Both models collected essentially the same price data. \hard{The base model spent its entire search budget on prices and, once the limit was reached, fabricated the two missing ratings; the hallucinated low score for Highland Park (83 instead of the true 88.91) exactly reverses the top-two ranking.} \good{Our model prioritized the scarcest evidence, namely the per-brand ratings, within the 7-search budget, obtained all four real ratings before computing the metric, and corrected an OCR-level misreading (``ORAN'' as Oban) using the region map. In aggregation tasks, data completeness matters more than last-step arithmetic.}
\end{casebox}

\begin{casebox}{BrowseComp-VL \;(\texttt{level1\_0080}): entity disambiguation with quoted-phrase queries}
\footnotesize
\begin{minipage}[t]{0.17\linewidth}
  \vspace{0pt}
  \centering
  \includegraphics[width=\linewidth]{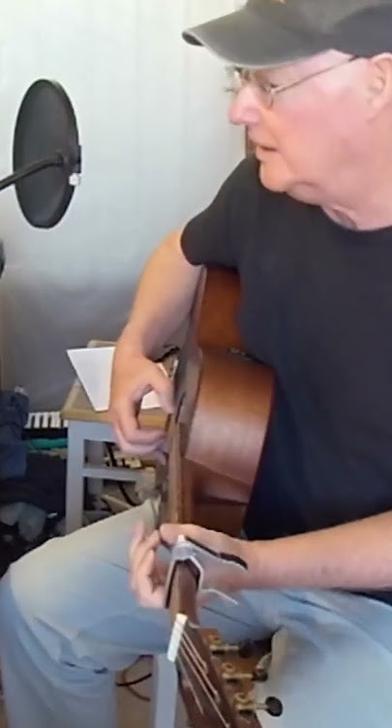}\\[-0.5mm]
  {\scriptsize Musician in a home studio}
\end{minipage}\hfill
\begin{minipage}[t]{0.80\linewidth}
  \vspace{0pt}
  \textbf{Task.} Was this musician's release of \emph{Smithereens} after \emph{Every Which Way}? Both titles correspond to \hard{highly confusable entities}: the band The Smithereens, Joji's \emph{Smithereens}, and Brian Davison's \emph{Every Which Way}.\\
  \textbf{Ground truth.} Yes (Dan Reeder: \emph{every which way} 2020 $\to$ \emph{Smithereens} 2024).
\end{minipage}

\vspace{1.2mm}
\begin{minipage}[t]{0.495\linewidth}
\vspace{0pt}
\begin{basepanel}
\scriptsize
\ttag{t0}\hspace{0.6em}\obs{web\_search(\texttt{"musician album Smithereens Every Which Way release order"})} $\to$ \obs{results are flooded by the three confusable entities}.

\ttag{t1}\hspace{0.6em}\obs{web\_search(\texttt{"`Every Which Way' album release date artist"})} $\to$ ``\hard{`Every Which Way' is an album by Brian Davison, released in 1970}.''

\ttag{t2}\hspace{0.6em}\obs{web\_search(\texttt{"Brian Davison Smithereens album"})} $\to$ \obs{no ``Smithereens'' by Davison}.

\ttag{t3--t5}\hspace{0.6em}\obs{web\_search(\texttt{"Brian Davison musician Every Which Way 1970"}), \dots, \texttt{Wikipedia discography}} $\to$ \hard{anchors on Brian Davison} (a drummer, while the pictured man plays guitar) and never revises the anchor.

\ttag{t6}\hspace{0.6em}\obs{visit(Wikipedia: Brian Davison)} $\to$ \obs{full discography, no ``Smithereens''}.

\ttag{t7--t8}\hspace{0.6em}\obs{web\_search(\texttt{"`Smithereens' album Brian Davison"}), \; web\_search(\texttt{"`Smithereens' album release date artist"})} $\to$ ```Smithereens' refers to a different band entirely.''

\ttag{t9 [answer]}\hspace{0.6em}``Brian Davison released `Every Which Way' in 1970; no album called `Smithereens'.'' $\Rightarrow$ \hard{No} ($\times$)
\end{basepanel}
\end{minipage}\hfill
\begin{minipage}[t]{0.495\linewidth}
\vspace{0pt}
\begin{ourspanel}
\scriptsize
\ttag{t0}\hspace{0.6em}\obs{web\_search(\texttt{"`Every Which Way' album Smithereens musician release after"})} $\to$ \obs{mixed signals: The Smithereens, Joji}.

\ttag{t1}\hspace{0.6em}\obs{web\_search(\texttt{"Brian Davison Every Which Way musician"})} $\to$ \obs{Davison formed Every Which Way after The Nice}.

\ttag{t2}\hspace{0.6em}\obs{web\_search(\texttt{"Brian Davison Smithereens release after Every Which Way"})} $\to$ \obs{no connection found}.

\ttag{t3--t4}\hspace{0.6em}\obs{web\_search(\texttt{"`The Smithereens' `Every Which Way' album song"}), \; web\_search(\texttt{"Joji `Every Which Way' album"})} $\to$ \obs{neither matches the pictured musician}.

\ttag{t5}\hspace{0.6em}\obs{image\_search(\texttt{"Brian Davison musician photo"})} $\to$ ``He was a \good{drummer, not a guitarist}. \good{The person in the image is playing a guitar}.'' \obs{(image search falsifies the anchor)}.

\ttag{t6}\hspace{0.6em}\obs{web\_search(\texttt{\good{"Smithereens" "Every Which Way" same artist}})} $\to$ \obs{``Smithereens, song and lyrics by \good{Dan Reeder}''} \obs{(the dual-phrase quoted query disambiguates in one step)}.

\ttag{t7}\hspace{0.6em}\obs{image\_search(\texttt{"Dan Reeder musician photo"})} $\to$ \obs{photos of Reeder with a guitar, matching the input image}.

\ttag{t8 [answer]}\hspace{0.6em}``\emph{every which way} (2020) $\to$ \emph{Smithereens} (2024): \good{YES}, `Smithereens' was released after `Every Which Way'.'' $\Rightarrow$ \good{Yes} (pass)
\end{ourspanel}
\end{minipage}

\vspace{1.2mm}
\textbf{What made the difference.} The question is a simple release-order comparison, and the only difficulty is \hard{resolving the entity}: the base model let an early, plausible result (Brian Davison, 1970) fix the anchor, then spent the remaining budget trying to force ``Smithereens'' onto the wrong artist. \good{Our model actively falsified the anchor with an image search (drummer versus guitarist), then used a dual-phrase quoted query that co-occurs both titles with one artist, which disambiguated in a single step.} The contrast shows that when the answer hinges on entity identity, \good{evidence-based anchor revision} is the decisive node, not more searching.
\end{casebox}

\begin{casebox}{MMSearch-Plus \;(\texttt{mmsearch\_plus\_0274}): cross-language intent formulation}
\footnotesize
\begin{minipage}[t]{0.17\linewidth}
  \vspace{0pt}
  \centering
  \includegraphics[width=\linewidth]{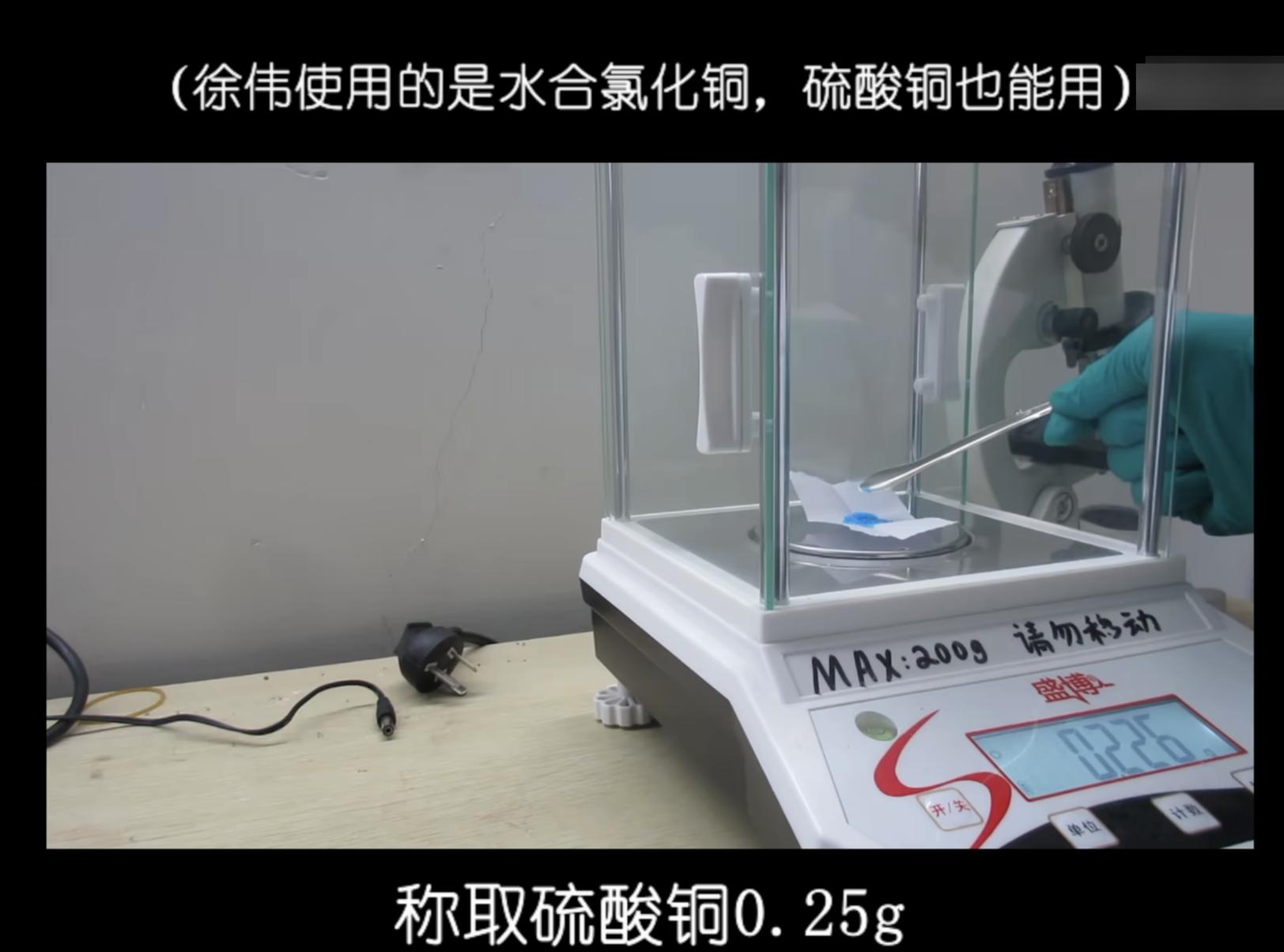}\\[-0.5mm]
  {\scriptsize Experiment records}
\end{minipage}\hfill
\begin{minipage}[t]{0.80\linewidth}
  \vspace{0pt}
  \textbf{Task.} Read the handwritten experiment records in the image (reagents \emph{copper chloride}, \emph{copper sulfate}, a dosage of \emph{0.25 g}, and the word \hard{disease}) and determine \hard{which disease this experiment targets}.\\
  \textbf{Ground truth.} Menkes disease (copper transporter ATP7A, a gene-therapy experiment by a specific researcher).
\end{minipage}

\vspace{1.2mm}
\begin{minipage}[t]{0.495\linewidth}
\vspace{0pt}
\begin{basepanel}
\scriptsize
\ttag{t0}\hspace{0.6em}\obs{web\_search(\texttt{"[researcher name] copper chloride copper sulfate experiment"})} \obs{(queried in Chinese)} $\to$ \obs{generic chemistry experiments}.

\ttag{t1 [answer]}\hspace{0.6em}``A chemistry experiment involving copper sulfate and copper chloride\dots \hard{taste experiments}'' $\Rightarrow$ \hard{wrong disease} ($\times$)
\end{basepanel}
\end{minipage}\hfill
\begin{minipage}[t]{0.495\linewidth}
\vspace{0pt}
\begin{ourspanel}
\scriptsize
\ttag{t0}\hspace{0.6em}\obs{web\_search(\texttt{"copper sulfate experiment targeting disease 0.25g"})} $\to$ \obs{generic; no specific disease}.

\ttag{t1}\hspace{0.6em}\obs{web\_search(\texttt{"`0.25g' copper sulfate disease experiment"})} $\to$ \obs{still ambiguous}.

\ttag{t2}\hspace{0.6em}\obs{web\_search(\texttt{"[researcher name] copper chloride copper sulfate experiment disease"})} \obs{(queried in Chinese, adding the intent word ``disease'')} $\to$ \obs{hits the researcher's gene-therapy records} $\Rightarrow$ \good{Menkes disease (ATP7A)}.

\ttag{t3 [answer]}\hspace{0.6em}``The experiment targets \good{Menkes disease}, a rare genetic disorder caused by mutations in \good{ATP7A}.'' $\Rightarrow$ \good{correct disease} (pass)
\end{ourspanel}
\end{minipage}

\vspace{1.2mm}
\textbf{What made the difference.} The deciding evidence lives in Chinese-language sources, so both models eventually switch to a Chinese query. \hard{The base model switched language but dropped the intent word ``disease'' and the dosage, and its query returned generic copper-chemistry pages, from which it guessed ``taste experiments''.} \good{Our model first exhausted English queries, then reformulated the Chinese query to preserve both the reagents and the intent constraint ``disease'', which is exactly the combination that retrieves the relevant records.} This case isolates \good{intent-preserving cross-language reformulation} as the critical node.
\end{casebox}

\begin{casebox}{MMSearchExam \;(\texttt{mmsearchexam\_0212}): visual clues fused into a single query}
\footnotesize
\begin{minipage}[t]{0.17\linewidth}
  \vspace{0pt}
  \centering
  \includegraphics[width=\linewidth]{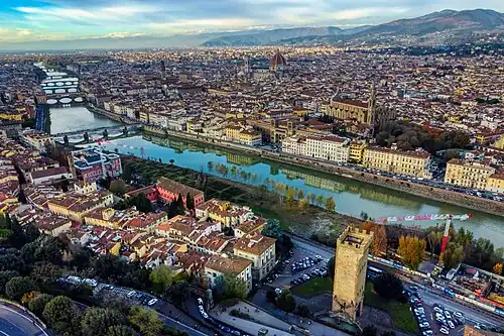}\\[-0.5mm]
  {\scriptsize De Chirico, \emph{The Song of Love}}
\end{minipage}\hfill
\begin{minipage}[t]{0.80\linewidth}
  \vspace{0pt}
  \textbf{Task.} Identify the painting (a \hard{Greek sculpted head}, a \hard{surgeon's glove}, a \hard{green ball}, and a \hard{locomotive on the horizon}), then state its \hard{original title in the language originating from the city shown in the image}.\\
  \textbf{Ground truth.} \emph{La canzone dell'amore} (Italian, from Florence).
\end{minipage}

\vspace{1.2mm}
\begin{minipage}[t]{0.495\linewidth}
\vspace{0pt}
\begin{basepanel}
\scriptsize
\ttag{t0}\hspace{0.6em}\obs{web\_search(\texttt{"painting Greek sculpted head surgeon's glove green ball locomotive"})} $\to$ \obs{identifies \emph{The Song of Love} (De Chirico)}.

\ttag{t1}\hspace{0.6em}\obs{web\_search(\texttt{"`The Song of Love' De Chirico 1914 original Italian title"})} $\to$ \obs{a search snippet reports the title as \hard{``Le Chant d'amour''} (French)}.

\ttag{t2 [answer]}\hspace{0.6em}``The original title in Italian\dots is \hard{Le Chant d'amour}.'' $\Rightarrow$ \hard{French title given} ($\times$)
\end{basepanel}
\end{minipage}\hfill
\begin{minipage}[t]{0.495\linewidth}
\vspace{0pt}
\begin{ourspanel}
\scriptsize
\ttag{t0}\hspace{0.6em}\obs{web\_search(\texttt{"painting Greek sculpted head surgeon's glove green ball locomotive"})} $\to$ \obs{identifies \emph{The Song of Love} (De Chirico, 1914)}.

\ttag{t1 [answer]}\hspace{0.6em}``The city shown is \good{Florence, Italy}; the language originating from Florence is \good{Italian}; the original Italian title is \good{\emph{La canzone dell'amore}}.'' $\Rightarrow$ \good{correct Italian title} (pass)
\end{ourspanel}
\end{minipage}

\vspace{1.2mm}
\textbf{What made the difference.} Both models identify the painting in one step; the only remaining question is the language constraint. \hard{The base model accepted a French snippet verbatim (``Le Chant d'amour'') without checking it against the city in the image, and answered in French.} \good{Our model re-derived the language from the image evidence (Florence, hence Italian) and answered with the Italian title, ignoring the misleading French phrasing.} The case shows that \good{constraint adherence, that is, matching the answer to the requested language, beats accepting the first retrieved phrase}, and the image evidence serves as the authority that overrides the search snippet.
\end{casebox}

\begin{casebox}{TIR-Bench \;(\texttt{tir\_bench\_0088}): quantitative reasoning instead of visual guessing}
\footnotesize
\begin{minipage}[t]{0.17\linewidth}
  \vspace{0pt}
  \centering
  \includegraphics[width=\linewidth]{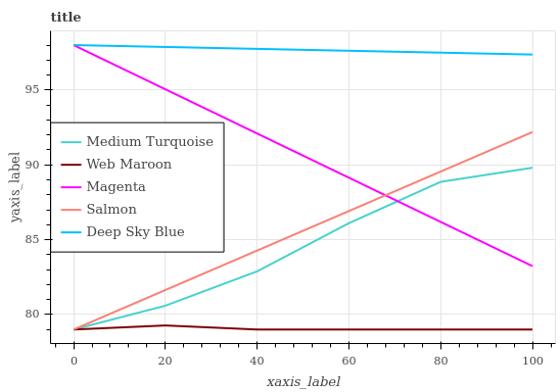}\\[-0.5mm]
  {\scriptsize Line chart with two curves}
\end{minipage}\hfill
\begin{minipage}[t]{0.80\linewidth}
  \vspace{0pt}
  \textbf{Task.} The chart plots a salmon-colored curve (rising from $y \approx 79$ to $y \approx 92$) and a magenta curve (falling from $y \approx 98$ to $y \approx 83$). Select the option that best matches the \hard{x-coordinate of the intersection}.\\
  \textbf{Ground truth.} C. 68.
\end{minipage}

\vspace{1.2mm}
\begin{minipage}[t]{0.495\linewidth}
\vspace{0pt}
\begin{basepanel}
\scriptsize
\ttag{t0 [answer]}\hspace{0.6em}``At $x=60$: Salmon $\approx 84$, Magenta $\approx 89$; at $x=65$: 85.5 and 87.5; at $x=66$: both $\approx 86$ \obs{(purely visual interpolation)}. This suggests the intersection is around $x=66$.'' $\Rightarrow$ \hard{B. 66} ($\times$)
\end{basepanel}
\end{minipage}\hfill
\begin{minipage}[t]{0.495\linewidth}
\vspace{0pt}
\begin{ourspanel}
\scriptsize
\ttag{t0 [answer]}\hspace{0.6em}``Read the two endpoints: Salmon $y=79\to 92$ (slope $+0.13$), Magenta $y=98\to 83$ (slope $-0.15$). Set \good{$98-0.15x = 79+0.13x$}; solve: $19=0.28x$, so \good{$x \approx 67.86 \approx 68$}.'' $\Rightarrow$ \good{C. 68} (pass)
\end{ourspanel}
\end{minipage}

\vspace{1.2mm}
\textbf{What made the difference.} Both models read the same endpoint values from the chart. \hard{The base model stopped at visual interpolation, which drifts toward 66, an error of 2 units.} \good{Our model turned the two straight-line approximations into a linear equation and solved it exactly, yielding 67.86, which selects option 68.} The case illustrates \good{quantitative verification over perceptual judgment} at the final decision node.
\end{casebox}


\begin{casebox}{VisBrowse \;(\texttt{visbrowse\_0142}): one query with all constraints vs.\ budget exhaustion}
\footnotesize
\begin{minipage}[t]{0.17\linewidth}
  \vspace{0pt}
  \centering
  \includegraphics[width=\linewidth]{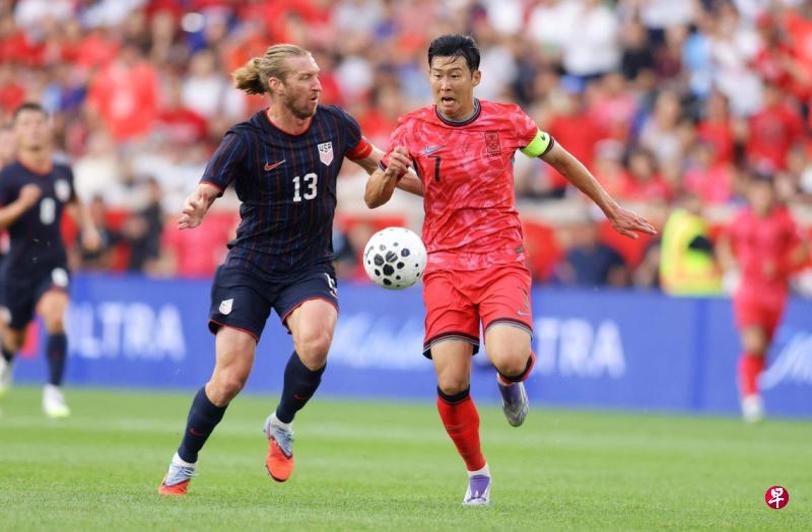}\\[-0.5mm]
  {\scriptsize Athlete with New Jeans}
\end{minipage}\hfill
\begin{minipage}[t]{0.80\linewidth}
  \vspace{0pt}
  \textbf{Task.} The athlete in the red shirt (Son Heung-min) took a photo with members of New Jeans. Identify the \hard{jersey number held by the girl to his right} in that photo.\\
  \textbf{Ground truth.} 6.
\end{minipage}

\vspace{1.2mm}
\begin{minipage}[t]{0.495\linewidth}
\vspace{0pt}
\begin{basepanel}
\scriptsize
\ttag{t0}\hspace{0.6em}\obs{web\_search(\texttt{"Son Heung-min New Jeans photo jersey number"})} $\to$ \obs{general pages}.

\ttag{t1--t3}\hspace{0.6em}\obs{web\_search(\texttt{"Son Heung-min New Jeans photo members"}), \; web\_search(\texttt{"Son Heung-min NewJeans photo jersey number member"}), \; visit(Instagram)} $\to$ \obs{no exact photo; visit fails} $[\ldots]$.

\ttag{t4--t13}\hspace{0.6em}\obs{web\_search(\texttt{"NewJeans Son Heung-min photo jersey number Minji\dots"}), \; \dots, \; web\_search(\texttt{"NewJeans Son Heung-min Kim Min-jae photo jersey"})} $\to$ \hard{repeats variants of the same under-specified query; the constraint ``girl to his right'' never appears} $[\ldots]$ \obs{(queries scroll through member names, 13/9/7, Nike)}.

\ttag{t14--t15}\hspace{0.6em}\obs{zoom / image\_search on an allkpop URL} $\to$ \obs{fails to load the photo}.

\ttag{t16--t19}\hspace{0.6em}\obs{web\_search(\texttt{"NewJeans Son Heung-min photo jersey number 13"}), \; \dots} $\to$ \obs{exhausts the search limit}.

\ttag{t20 [answer]}\hspace{0.6em}``Error: Reached max turns without a definitive answer.'' $\Rightarrow$ \hard{no answer} ($\times$)
\end{basepanel}
\end{minipage}\hfill
\begin{minipage}[t]{0.495\linewidth}
\vspace{0pt}
\begin{ourspanel}
\scriptsize
\ttag{t0}\hspace{0.6em}\obs{web\_search(\texttt{\good{"Son Heung-min New Jeans photo jersey number girl to his right"}})} $\to$ \obs{first result states the exact fact: the girl to his right holds \good{number 6}}.

\ttag{t1 [answer]}\hspace{0.6em}``The search result says \good{`Son definitely took the girl hold number 6 home'}\dots the girl to his right is holding jersey number \good{6}.'' $\Rightarrow$ \good{6} (pass)
\end{ourspanel}
\end{minipage}

\vspace{1.2mm}
\textbf{What made the difference.} The answer exists in the top search result, but only for a query that encodes \emph{all} constraints. \hard{The base model spent all 21 turns on under-specified query variants (adding member names, jersey numbers 13/9/7, the Nike brand) and on failing zoom/visit calls; because the positional constraint ``girl to his right'' never entered the query, the searchable evidence was never retrieved.} \good{Our model composed a single query that bundles the person, the group, the object, and the positional constraint, which returns the answer immediately.} This is the cleanest illustration of \good{constraint-complete query construction} as the decisive node, and of \good{search-budget allocation} on the first attempt.
\end{casebox}

\begin{casebox}{VisualToolBench \;(\texttt{biology\_singleturn\_0013}): image evidence plus concept verification}
\footnotesize
\begin{minipage}[t]{0.17\linewidth}
  \vspace{0pt}
  \centering
  \includegraphics[width=\linewidth]{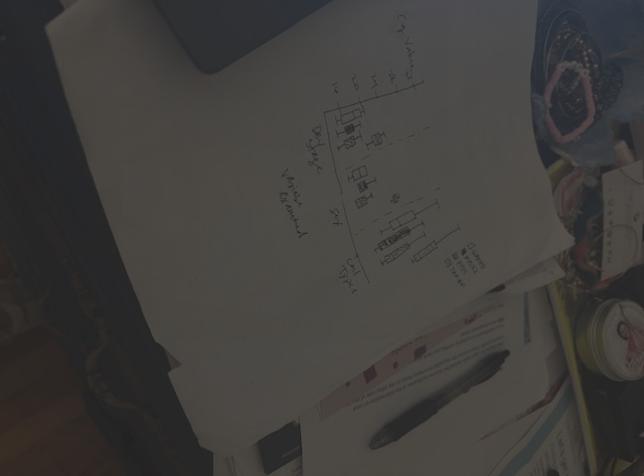}\\[-0.5mm]
  {\scriptsize RT-qPCR box plot}
\end{minipage}\hfill
\begin{minipage}[t]{0.80\linewidth}
  \vspace{0pt}
  \textbf{Task.} From the RT-qPCR box plot, name the gene with the \hard{lowest median Cq value} across developmental stages, and \hard{explain why having the lowest median Cq value does not automatically make it the best reference gene} for normalization.\\
  \textbf{Ground truth.} RPS5; because a good reference gene must be \emph{stably} expressed (low variability), not merely highly expressed.
\end{minipage}

\vspace{1.2mm}
\begin{minipage}[t]{0.495\linewidth}
\vspace{0pt}
\begin{basepanel}
\scriptsize
\ttag{t0--t6}\hspace{0.6em}\obs{zoom / image reads} \obs{(repeatedly crops the same chart, reads x-axis labels, box plots, legend)} $[\ldots]$ \obs{(never searches; the required knowledge of reference-gene criteria is not in the image)}.

\ttag{t7 [answer]}\hspace{0.6em}``The image quality is insufficient to clearly read all the labels; I cannot determine the gene.'' $\Rightarrow$ \hard{no answer} ($\times$)
\end{basepanel}
\end{minipage}\hfill
\begin{minipage}[t]{0.495\linewidth}
\vspace{0pt}
\begin{ourspanel}
\scriptsize
\ttag{t0}\hspace{0.6em}\obs{web\_search(\texttt{"lowest median Cq value gene best reference gene normalization"})} $\to$ \obs{definitions of median Cq and reference-gene selection}.

\ttag{t1}\hspace{0.6em}\obs{web\_search(\texttt{"`lowest median Cq value' reference gene why not best"})} $\to$ \obs{\good{the stability criterion} (low variability across conditions) is retrieved}.

\ttag{t2 [answer]}\hspace{0.6em}``\good{RPS5} has the lowest median Cq value; however, the lowest median Cq alone does not imply the best reference gene, because \good{a reference gene must be stably expressed} (tight clustering, low variability) across all conditions.'' $\Rightarrow$ \good{both parts answered} (pass)
\end{ourspanel}
\end{minipage}

\vspace{1.2mm}
\textbf{What made the difference.} The factual part (which gene) lives in the image, but the conceptual part (why not automatically the best) lives outside it. \hard{The base model never left the zoom tool, and after several failed crops it gave up with ``image quality insufficient'', although it had already located the legend.} \good{Our model read the chart, then used two targeted searches to both confirm the gene and retrieve the stability criterion, so it answered the two-part question completely.} This case shows that \good{mixing image grounding with external verification} is required whenever a question couples a visual fact with domain knowledge.
\end{casebox}